\documentclass[sigconf]{acmart}

\usepackage{amsmath,amsfonts,bm}
\usepackage{graphicx}
\usepackage{fancyhdr}
\usepackage{titlesec}

\def\eqref#1{equation~\ref{#1}}

\def\1{\bm{1}}

\def\va{{\bm{a}}}

\def\vh{{\bm{h}}}

\def\mA{{\bm{A}}}

\def\mH{{\bm{H}}}

\def\mM{{\bm{M}}}

\def\mW{{\bm{W}}}
\def\mX{{\bm{X}}}

\DeclareMathAlphabet{\mathsfit}{\encodingdefault}{\sfdefault}{m}{sl}
\SetMathAlphabet{\mathsfit}{bold}{\encodingdefault}{\sfdefault}{bx}{n}

\def\gE{{\mathcal{E}}}

\def\gG{{\mathcal{G}}}

\def\gL{{\mathcal{L}}}
\def\gM{{\mathcal{M}}}
\def\gN{{\mathcal{N}}}

\def\gV{{\mathcal{V}}}

\def\sR{{\mathbb{R}}}

\newcommand{\softmax}{\mathrm{softmax}}

\usepackage[linesnumbered,ruled, vlined]{algorithm2e}
\usepackage{algorithmic}
\usepackage{algorithmic}
\usepackage{subfigure} 
\usepackage{multirow}

\AtBeginDocument{%
  \providecommand\BibTeX{{%
    \normalfont B\kern-0.5em{\scshape i\kern-0.25em b}\kern-0.8em\TeX}}}

\newcommand{\eat}[1]{}

\usepackage{xcolor,comment}
\usepackage[normalem]{ulem}

\setcopyright{acmcopyright}
\copyrightyear{2018}
\acmYear{2018}
\acmDOI{XXXXXXX.XXXXXXX}

\acmConference[Conference acronym 'XX]{Make sure to enter the correct
  conference title from your rights confirmation emai}{June 03--05,
  2018}{Woodstock, NY}
\acmPrice{15.00}
\acmISBN{978-1-4503-XXXX-X/18/06}

\begin{document}

\title{DyExplainer: Explainable Dynamic Graph Neural Networks}




\author{Tianchun Wang$^{1}$, Dongsheng Luo$^{2}$, Wei Cheng$^{3}$, Haifeng Chen$^{3}$, Xiang Zhang$^{1}$}
\affiliation{%
  \institution{$^1$Pennsylvania State University, $^2$Florida International University, $^3$NEC Labs America
  \country{}}
}
\email{
{tkw5356,xzz89}@psu.edu, dluo@fiu.edu, {weicheng, haifeng}@nec-labs.com
}







\renewcommand{\shortauthors}{Trovato and Tobin, et al.}

\begin{abstract}
  Graph Neural Networks (GNNs) resurge as a trending research subject owing to their impressive ability to capture representations from graph-structured data. However, the black-box nature of GNNs presents a significant challenge in terms of comprehending and trusting these models, thereby limiting their practical applications in mission-critical scenarios. Although there has been substantial progress in the field of explaining GNNs in recent years, the majority of these studies are centered on static graphs, leaving the explanation of dynamic GNNs largely unexplored. Dynamic GNNs, with their ever-evolving graph structures, pose a unique challenge and require additional efforts to effectively capture temporal dependencies and structural relationships. To address this challenge, we present DyExplainer, a novel approach to explaining dynamic GNNs on the fly. DyExplainer trains a dynamic GNN backbone to extract representations of the graph at each snapshot, while simultaneously exploring structural relationships and temporal dependencies through a sparse attention technique. To preserve the desired properties of the explanation, such as structural consistency and temporal continuity, we augment our approach with contrastive learning techniques to provide priori-guided regularization. To model longer-term temporal dependencies, we develop a buffer-based live-updating scheme for training. The results of our extensive experiments on various datasets demonstrate the superiority of DyExplainer, not only providing faithful explainability of the model predictions but also significantly improving the model prediction accuracy, as evidenced in the link prediction task.
\end{abstract}

\maketitle
\section{Introduction}
\label{sec:intro}

The advent of Graph Neural Networks (GNNs) has caused a veritable revolution in the field, and has been embraced with great enthusiasm due to their demonstrated efficacy in a variety of applications, ranging from node classification~\cite{kipf2016semi} and link prediction~\cite{zhang2018link}, to graph clustering~\cite{liu2023deep} and recommender systems~\cite{wu2022graph}.
However, these models 
are usually treated as black boxes, and their predictions lack understanding and explanations, thus preventing them to produce reliable solutions. Therefore, the study of the explainability of GNNs is in need. Recent research on GNN explainability mainly focuses on explaining the model predictions. 
However, the labyrinthine nature of GNNs has often resulted in their predictions being shrouded in mystery and lacking transparency. Consequently, the trustworthiness of their solutions is frequently called into question. To rectify this, the exploration and understanding of the explainability of GNNs have become imperative. The current research landscape in this arena primarily centers on demystifying the underlying mechanisms of GNN predictions, utilizing methods such as gradient-based techniques~\cite{baldassarre2019explainability,pope2019explainability}, mask-based approaches~\cite{ying2019gnnexplainer,luo2020parameterized,yuan2021explainability}, and surrogate models~\cite{vu2020pgm} to shed light on the reasoning~\cite{Liangke_Survey} behind them.
These techniques endeavor to bring greater transparency to the predictions of GNNs and to provide a deeper insight into the thought process of the model.

The advancement in explainability in static GNNs has been substantial, yet the same cannot be said for dynamic GNNs. Despite this, dynamic GNNs have gained widespread use in real-world applications, as they are capable of handling graphs that change over time. Elevating the level of explainability in dynamic GNNs is of utmost importance as it can foster greater human trust in the model's predictions. However, the dynamic and evolving nature of graphs presents several challenges. Firstly, learning in dynamic graphs involves taking into consideration not only the topology and node attributes at each time point but also the temporal dynamics, making it a formidable task to present explanations in a manner that is easily understood by humans. Secondly, the consecutive order of dynamic graphs imposes unique constraints on the continuity of explanations. Lastly, the underlying patterns in dynamic graphs may also evolve with changes in node features and topologies, making it a challenge to find a balance between the continuity and evolution of explanations.

\begin{figure*}[ht]
    \centering
\includegraphics[width=0.74\textwidth]{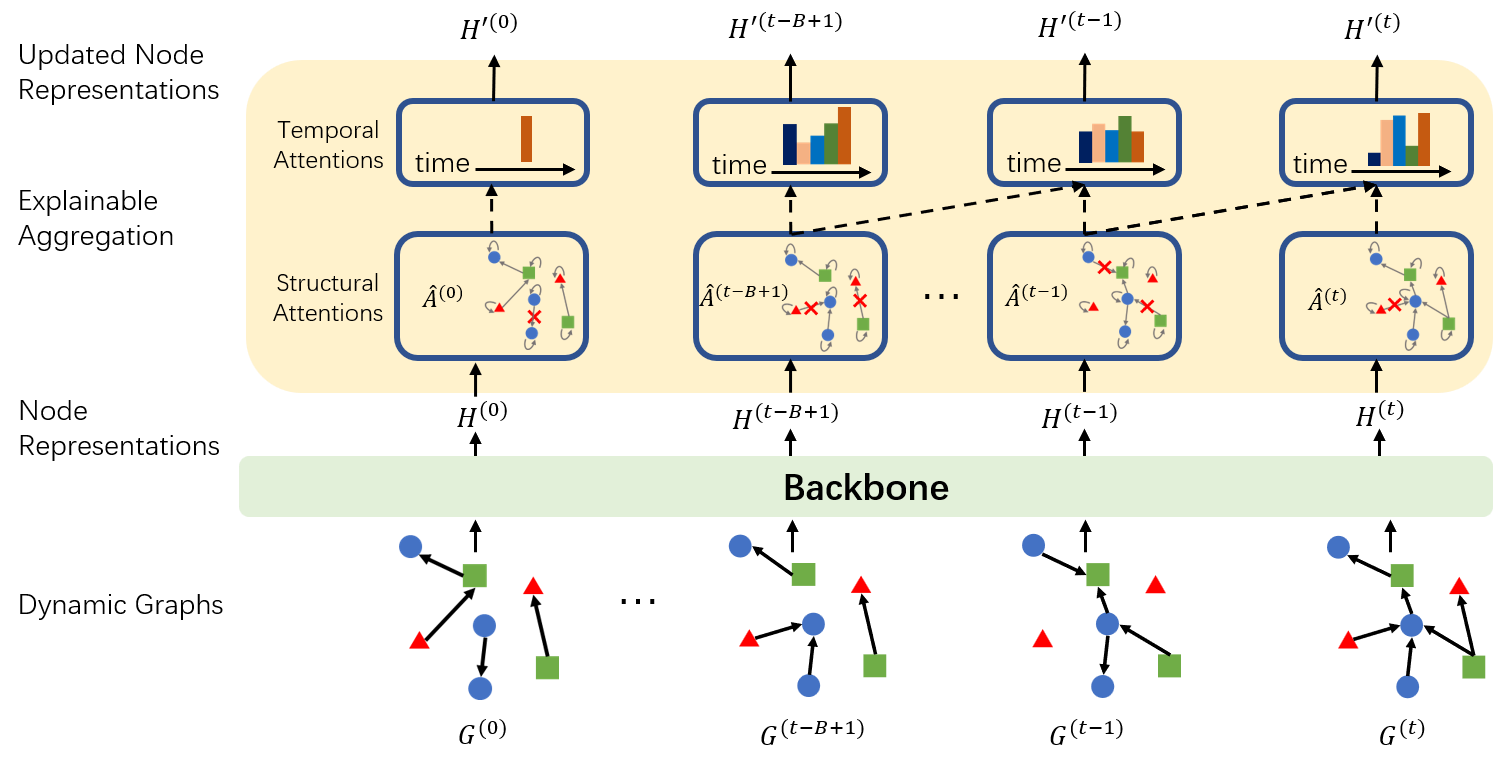}
    \caption{DyExplainer for dynamic graphs. $\mH^{(t)}$ is the node representation for snapshot $G^{(t)}$ after backbone module. $\mH'^{(t)}$ is the updated node representation after explainable module. $\widehat{\mA}^{(t)}$ is the structural attention for snapshot $G^{(t)}$.}
    \label{fig:aim3.1:framework}
    \vspace{-3mm}
\end{figure*}

To handle these challenges, in this paper, we propose DyExplainer, a dynamic explainable mechanism for dynamically interpreting GNNs. DyExplainer constitutes a high-level, generalizable explainer that imparts insightful and comprehensible explanations through the utilization of graph patterns, thereby providing a deeper understanding of predictions made by diverse dynamic GNNs.
The proposed approach incorporates pooling operations~\cite{ying2018hierarchical,ma2019graph} on nodes, thereby yielding graph-level embeddings that are encoder-agnostic and afford a remarkable degree of flexibility to a broad spectrum of dynamic encoders serving as the backbone. Furthermore, the introduction of the explainer module into the encoder incurs only a minimal overhead, making it highly efficient for large-scale backbone networks. Specifically, DyExplainer trains a dynamic GNN model to produce node embeddings at each snapshot, where the explainer module comprises two attention components: structural attention and temporal attention. The former leverages the pivotal relationships between nodes within the graph at each snapshot to inform the node representations, while the latter accounts for the temporal dependencies between representations generated by long-term snapshots. Despite the proliferation of dynamic GNNs~\cite{you2022roland,rossi2020temporal}, most of these approaches deduce the representation at each snapshot $t$ solely based on the embedding at the previous snapshot $(t-1)$, thereby making it challenging to fully comprehend the intricate, long-term temporal dependencies between snapshots in real-world applications. For instance, social network connections between individuals, groups, and communities may be subject to change over time, influenced by a multitude of factors, such as geographical distance, personal life stage, and evolving interests. To encompass longer-term temporal dependencies, we have devised a buffer-based, live-updating scheme with temporal attention. Specifically, the temporal attention aggregates the node embeddings from the preceding $B$ snapshots stored in a buffer. In this regard, prior works on dynamic GNNs are reduced to a special case of the DyExplainer, where $B=1$. The proposed framework constitutes a generalization that encompasses all recent graph learning methods for dynamic graphs. For example, by relinquishing explainability and incorporating the Markov chain property for temporal evolution, our framework degenerates to ROLAND~\cite{you2022roland}. Additionally, by restricting temporal aggregations to only the final layer, our framework degenerates to a common approach in which a sequence model, such as GRU~\cite{cho2014properties}, is placed on top of GNNs~\cite{peng2020spatial,wang2020traffic,yu2017spatio}.

The dual sparse attention components of DyExplainer serve a trifold function. Firstly, they impart incisive explanations for the model's predictions in downstream tasks. Secondly, they serve as a sparse regularizer, enhancing the learning process. As research in the field of ranshomon set theory~\cite{ranshomon1,ranshomon2} demonstrates, sparse and interpretable models possess a higher degree of robustness and better generalization performance, as outlined in Sec.~\ref{sec:ablation}. Lastly, the attention components allow for the augmentation of the approach with priori-guided regularization, preserving the desirable properties of the explanation, such as structural consistency and temporal continuity. Structural consistency pertains to the consistency of node embeddings between connected nodes in a single snapshot, while temporal continuity enforces smoothness constraints on the temporal attention between closely spaced snapshots, guided by pre-established human priors. To achieve more lucid explanations, we employ the use of contrastive learning techniques, treating connected pairs as positive examples and unconnected pairs as negative examples for consistency regularization, and recent snapshots as positive examples and distant historical snapshots as negative examples for continuity regularization. Overall, our contributions are summarized as follows:
\vspace{-4mm}
\begin{itemize}
\item We put forward the problem of explainable dynamic GNNs and propose a general DyExplainer to tackle it. As far as we know, it is the first work solving this problem.

\item DyExplainer seamlessly integrates the modeling of both structural relationships and long-term dependencies via sparse attentions. As a result, it is capable of providing real-time explanations for both the structural and temporal factors that influence the model's predictions. This innovative explaining module has been designed to be encoder-agnostic, thereby affording flexibility to a range of dynamic GNN backbones. Its implementation requires minimal overhead, as it only entails adding a lightweight parameterization to the encoder for its explanation modules. This makes DyExplainer highly efficient even for large backbone networks.

\item  We propose two contrastive regularizations to provide consistency and continuity explanations. Our approach to augmenting desired properties in the explanation is a fresh contribution to the field and may be of independent interest.

\item Extensive experiments on various datasets demonstrate the superiority of DyExplainer, not only providing faithful explainability of the model predictions but also significantly improving the model prediction accuracy, as evidenced in the link prediction
task.
\end{itemize}
\begin{figure}
    \centering
\includegraphics[width=0.25\textwidth]{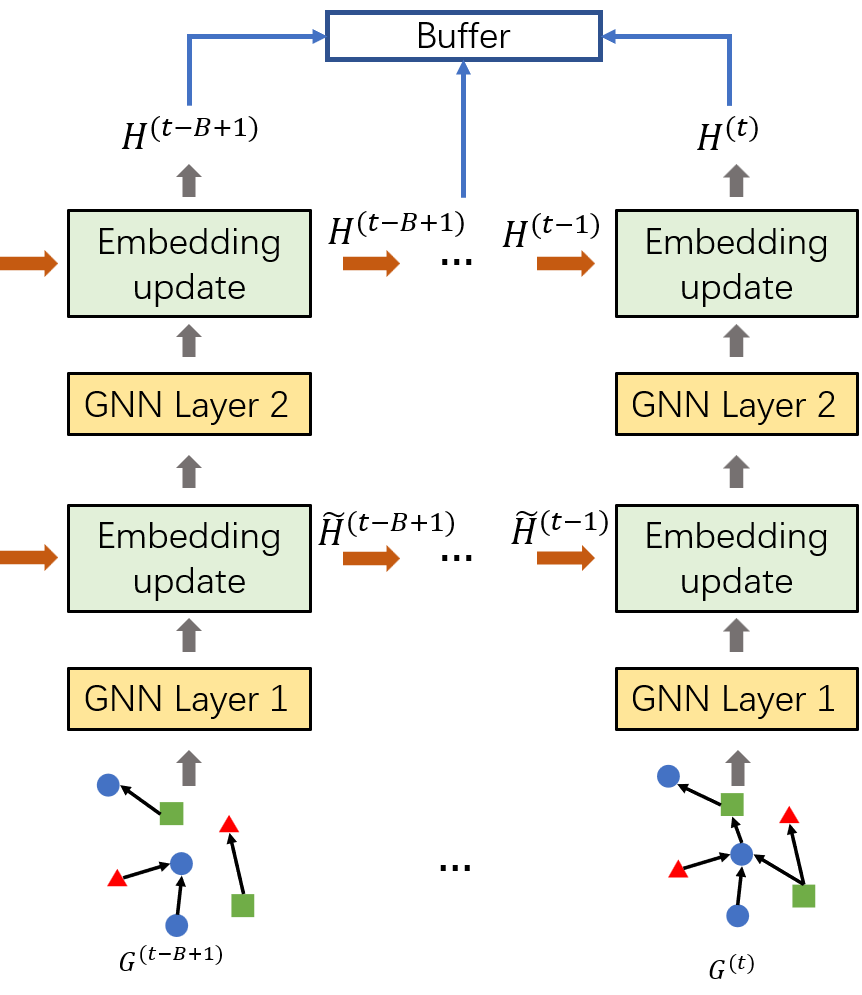}
 \vspace{-1em}
    \caption{Backbone encoder architecture.}
    \label{fig:backbone}
 \vspace{-2em}
\end{figure}

\section{Problem Definition}
We formulate a dynamic graph as a series of $T$ attributed graphs as $\gG=(G^{(1)},...,G^{(T)})$, where each graph $G^{(t)}=(\gV, \gE^{(t)}, \mX^{(t)})$ is a snapshot at time step $t$. $\gV = \{v_1,v2,...v_N\}$ is the set of $N$ nodes shared by all snapshots and $\gE^{(t)} \in \gV \times \gV $ is the set of edges of a graph $G^{(t)}$. $\mX^{(t)} \in \sR^{N\times D}$ is the node feature matrix at snapshot $t$, where $D$ is the number of dimensions of node features. The topology and node features of each snapshot are dynamically changing over time.
We aim to learn an explainable dynamic GNN model $f$ from the long-term snapshots defined as:
\begin{align}
\widehat{\mA}^{(t-B+1)},...,\widehat{\mA}^{(t)}, \mH'^{(t+1)} \leftarrow f\left(G^{(t-B+1)}, ..., G^{(t)}\right),
\end{align}
where $\widehat{\mA}^{(t-B +1)},...,\widehat{\mA}^{(t)}$ are explanations we are interested and $\mH'^{(t+1)}$ is the future node representation predicted for the downstream tasks. In our study, we examine its application for link prediction. It's worth noting that our framework can also easily support other downstream tasks.

\section{Method}
\subsection{Encoder Backbone}
There are different approaches to designing dynamic GNNs. As a plug-and-use framework, the proposed DyExplainer is flexible to the choice of dynamic GNNs as the backbone. In this work, we adopt a state-of-the-art method ROLAND~\cite{you2022roland} as the backbone due to its powerful expressiveness and impressive empirical performances in real-life datasets.  Specifically, it hierarchically updates the node embedding to obtain the  $\mH^{(t)}$ at snapshot $t$. 
With ROLAND as the backbone, the architecture of the encoder is shown in Figure~\ref{fig:backbone}. 
We put the node embedding inferred by the backbone to a buffer of size $B$ for the learning of our explainer module. Formally, we denote the node embeddings in the buffer as $\{\mH^{(t-B+1)},...,\mH^{(t)}\}$. 


\subsection{Explainable Aggregations}
The node embedding at snapshot $t$ is denoted by $\mH^{(t)}=\{\vh_1,...,\vh_N\}$, $\vh_i \in \sR^F$, where $N$ is the number of nodes, and $F$ is the feature dimension of the node embedding. 
\subsubsection{Structural Aggregation}
\label{sec:str_agg}
On dynamic graphs, each snapshot has its unique topology information. An effective explanation, therefore, should highlight the crucial structural components at a given snapshot that significantly contribute to the model's prediction. To this end, inspired by the idea in~\cite{velivckovic2017graph}, we propose structural attention to aggregate the weighted neighborhoods at each time step. Formally, we have
\begin{align}
\label{eq:struct_omega}
&\omega^{(s,t)}_{ij}=\text{LeakyReLU}\Big(\va^{(s)T}\Big[\mW^{(s)}\vh^{(t)}_i||\mW^{(s)}\vh^{(t)}_j\Big]\Big),    \\
\label{eq:struct_att}
&\widehat{\mA}^{(t)}_{ij}=\softmax_j\Big(\omega^{(s,t)}_{ij}\Big)=\frac{\exp{\Big(\omega^{(s,t)}_{ij}}\Big)}{\sum_{k\in \gN^{(t)}_i}\exp{\Big(\omega^{(s,t)}_{ik}\Big)}},
\end{align}
where $\widehat{\mA}^{(t)}$ is the structural attention at time step $t$, $\mW^{(s)}$ is a weight matrix in a shared linear transformation, $\va^{(s)}$ is the weight vector in the single-layer network, $\mathcal{N}^{(t)}_i$ is the set of neighbors of node $i$ at time step $t$, and $||$ indicates the concatenation operation.

However, the utilization of weights in traditional attention models may pose a challenge in complex dynamic graph environments, particularly with regard to explanation. Explanations in these settings are often derived by imposing a threshold and disregarding insignificant attention weights. This approach, however, fails to account for the cumulative impact of numerous small but non-zero weights, which can be substantial. Moreover, the non-exclusivity of attention weights raises questions about their accuracy in reflecting the true underlying importance~\cite{velivckovic2017graph}. To address this issue and equip structural attention with better explainability, we design a hard attention mechanism to alleviate the effects of small attention coefficients. The basic idea uses a prior work on differentiable sampling~\citep{maddison2016concrete,jang2016categorical}, which states that the random variable
\begin{equation}\label{eq:gumbel}
e=\sigma\bigg(\Big(\log\epsilon-\log\big(1-\epsilon\big)+\omega\Big)/\tau\bigg)\quad\text{where}\quad\epsilon \sim \text{Uniform} \big(0, 1\big),
\end{equation}
where $\sigma(\cdot)$ is the sigmoid function and $\tau$ is the temperature controlling the approximation.
Equation~\ref{eq:gumbel} follows a distribution that converges to a Bernoulli distribution with success probability $p=(1+e^{-\omega})^{-1}$ as $\tau>0$ tends to zero. Hence, if we parameterize $\omega$ and specify that the presence of an edge between a pair of nodes has probability $p$, then using $e$ computed from~\eqref{eq:gumbel} to fill the corresponding entry of $\widehat{\mA}$ will produce a matrix $\widehat{\mA}$ that is close to binary. We use this matrix as the hard attention with the hope of obtaining a better explanation due to the dropping of small attention weights. Moreover, because~\eqref{eq:gumbel} is differentiable with respect to $\omega$, we can train the parameters of $\omega$ like in a usual gradient-based training. In the structural attention, we have the parameterized $\omega^{(s,t)}_{ij}$ in Equation~\ref{eq:struct_omega}, thus we get
\[
\widetilde{e}_{ij}^{(s,t)}=\sigma\bigg(\Big(\log\epsilon-\log\big(1-\epsilon\big)+\omega_{ij}^{(s,t)}\Big)/\tau\bigg)
\quad\text{where}\quad
\epsilon \sim \text{Uniform} \big(0, 1\big),
\]
which returns an approximate Bernoulli sample for the edge $(i,j)$. When $\tau$ is not sufficiently close to zero, this sample may not be close enough to binary, and in particular, it is strictly nonzero. The rationality of such an approximation is that with temperature $\tau >0 $, the gradient $\frac{\partial \widetilde{e}_{ij}^{(s,t)}}{\partial \omega_{ij}^{(s,t)}}$ is well-defined. The output of the binary concrete distribution is in the range of (0,1). To further alleviate the effects of small values by encouraging them to be exactly zero, we propose a ``stretching and clipping'' technique in the hard attention mechanism. 
To explicitly zero out an edge, we follow~\cite{louizos2017learning} and introduce two parameters, $\gamma < 0$ and $\xi > 1$, to remove small values of $\widetilde{e}_{ij}^{(s,t)}$given by
\[
\widehat{\mA}_{ij}^{(t)} = \min\Big(1, \max\big(e_{ij}^{(s,t)}, 0\big) \Big)
\quad\text{where}\quad
e_{ij}^{(s,t)} = \widetilde{e}_{ij}^{(s,t)} (\xi - \gamma) + \gamma.
\]
The structural attention $\widehat{\mA}^{(t)}$ does not insert new edges to the graph (i.e., when $(i,j)\notin\gE^{(t)}$, $\widehat{\mA}_{ij}^{(t)} = 0$), but only removes (denoises) some edges $(i,j)$ originally in $\gE^{(t)}$. Then, we obtain the node embedding $\widetilde{\mH}^{(t)} \in \sR^{N\times F'}$ at the time step $t$ after the structural aggregation with each row given by
\begin{equation}
\label{eq:struc_emb}
\widetilde{\vh}^{(t)}_i=\sigma\Big(\sum_{j \in \gN_i^{(t)}}\widehat{\mA}_{ij}^{(t)} \mW^{(s)} \vh_j^{(t)}\Big).
\end{equation}

\subsubsection{Temporal Aggregation}
\label{sec:temp_agg}
In light of the dynamic and evolving nature of node features and inter-node relationships over time, temporal dependency holds paramount importance in the modeling of dynamic graphs. Existing methods either adopt RNN architectures, such as GRU~\cite{cho2014properties} and LSTM~\cite{hochreiter1997long} or assume the Markow chain property~\cite{you2022roland} to capture the temporal dependencies in dynamic graphs.  As shown in~\cite{vaswani2017attention}, despite their utility, these methods are insufficient in capturing long-range dependencies, thereby hindering their ability to generalize and model previously unseen graphs. To overcome this limitation, we propose a solution that leverages an attention-based temporal aggregation mechanism to adaptively integrate node embeddings from distant snapshots. This is achieved through the utilization of a buffer-dependent temporal mask, which serves as a temporal topology to guide the aggregation process. 

In Equation~\ref{eq:struc_emb}, the structural attention provides the node embedding $\widetilde{\mH}^{(t)} \in \sR^{N\times F'}$ for each snapshot $t$. Therefore, we have a set of node embedding $\{\widetilde{\mH}^{(t-B+1)},...,\widetilde{\mH}^{(t)}\}$. Concatenating them to a 3-dimensional tensor and take transpose, for each node $i$, we have a buffer-dependent node embedding $\widehat{\mH}^{(i)} \in \sR^{B\times F'}$, $\widehat{\mH}^{(i)}=\{\widehat{\vh}_{t-B+1},...,\widehat{\vh}_t\}$, $\widehat{\vh}_t \in \sR^{F'}$.
We propose the temporal attention given by
\begin{align}
\label{eq:temp_omega}
&\omega^{(i)}_{t_k, t_j}=\text{LeakyReLU}\Big(\va^T\Big[\mW\vh^{(i)}_{t_k}||\mW \vh^{(i)}_{t_j}\Big]\Big),    \\
\label{eq:temp_att}
&\widetilde{\mA}^{(i)}_{t_k, t_j}=\softmax_{t_j}\Big(\omega^{(i)}_{t_k,t_j}\Big)=\frac{\exp{\Big(\omega^{(i)}_{t_k,t_j}}\Big)}{{\sum_{t_p\in \gM_{t_k}}}\exp{\Big(\omega^{(s,t)}_{t_k, t_p}\Big)}},
\end{align}
where $\widetilde{\mA}^{(i)}$ is the temporal attention for node $i$, $\mW$ is a weight matrix for linear transformation, $\va$ is a weight vector for single-layer network, $\gM_{t_k}$ is the time steps that has element 1 in the temporal mask. The values in $\widetilde{\mA}^{(i)} \in \sR^{B\times B}$ indicate the importance of relations between the embedding for node $i$ at the past snapshots. We compute Equation~\ref{eq:temp_omega} and \ref{eq:temp_att} in batch for acceleration due to 
the graphs usually have large numbers of nodes. Then, after the temporal attention, we obtain the node embedding $\mH^{'(i)} \in \sR^{B\times K}$ for each node $i$ for all the B snapshots in the buffer, with each row given by
\begin{equation}
    \label{eq:temp_emb}
\vh^{'(i)}_{t_k}=\sigma\Big(\sum_{t_j\in \mM_{t_k}}\widetilde{\mA}^{(i)}_{t_k,t_j}\mW\vh_{t_j}^{(i)}\Big).
\end{equation}
Therefore, we have the embedding of node $i$ at time $t$ is $\vh^{'(i)}_{t} \in \sR^{K}$, and the embedding of all nodes results in $\mH^{'(t)}\in\sR^{N\times K}$, $\mH^{'(t)}=\{\vh^{'(0)}_{t},...,\vh^{'(N-1)}_{t}\}$.

\subsection{Regularizations}
The framework of DyExplainer is flexible with various regularization terms to preserve desired
properties on the explanation. Inspired by the graph contrastive learning that makes the node representations more discriminative to capture different types of node-level similarity, we propose a structural consistency and a continuity consistency. 
We now discuss the regularization terms as well as their principles.
\subsubsection{Consistency Regularization}
Inspired by the homophily nature of graph-structured data~\cite{mcpherson2001birds}, we propose a topology-wise regularization to encourage consistent explanations of the connected nodes in a graph. Specifically, on the graph $G^{(t)}=\{\gV, \gE^{(t)}\}$, for a node $i \in \gV$, we sample a positive pair $(i,p)$, the edge $e_{i,p} \in \gE^{(t)}$. We sample unconnected pairs $(i,j)$ such that $e_{i,j} \notin \gE^{(t)}$ to form a set of non-negative samples $\bar{\mathcal{N}}_i$, then we propose the consistency regularization for the structural attention $\widehat{\mA}^{(t)}$ as
\begin{equation}
\label{eq:consistency_reg}
    \mathcal{L}_{cons} = -\log \frac{\exp\bigg(\text{sim}\Big(\widehat{\mA}^{(t)}_i,\widehat{\mA}^{(t)}_p\Big)\bigg)}{\exp\bigg(\text{sim}\Big(\widehat{\mA}^{(t)}_i,\widehat{\mA}^{(t)}_p\Big)\bigg)+\sum_{j\in \bar{\mathcal{N}}_i} \exp\bigg(\text{sim}\Big(\widehat{\mA}^{(t)}_i,\widehat{\mA}^{(t)}_j\Big)\bigg)}.
\end{equation}
Note that the computation of Equation~\ref{eq:consistency_reg} is very time-consuming for graphs with large numbers of nodes and edges. In practice, we select some anchors to compute the $\mathcal{L}_{cons}$.
\subsubsection{Continuity Regularization}
As suggested in~\cite{nauta2022anecdotal}, preserving continuity ensures the robustness of explanations that small variations applied to the input, for which the model prediction is nearly unchanged, will not lead to large differences in the explanation.  In addition, continuity benefits generalizability beyond a particular input instance. Based on the practical principle, in dynamic graphs, we aim to maintain a consistent explanation of snapshots even as the graph structure evolves. Inspired by the idea in~\cite{tonekaboni2021unsupervised} that two close subsequences are considered as a positive pair while the ones with large distances are the negatives, we propose a continuity regularization.

For a snapshot $G^{t}$, a positive pair $(G^{t}, G^{p})$ is sampled from the sliding window in the temporal mask. The set of non-negative samples  $\bar{\mathcal{N}}_t$ is historical snapshots that are not in the buffer. Then, the continuity regularization for each node $i$ is given by
\begin{equation}
\label{eq:continuity_reg}
    \mathcal{L}_{cont} = -\log \frac{\exp\bigg(\text{sim}\Big(\widetilde{\mA}^{(t)}_i,\widetilde{\mA}^{(p)}_i\Big)\bigg)}{\exp\bigg(\text{sim}\Big(\widetilde{\mA}^{(t)}_i,\widetilde{\mA}^{(p)}_i\Big)\bigg)+\sum_{k\in \bar{\mathcal{N}}_t} \exp\bigg(\text{sim}\Big(\widetilde{\mA}^{(t)}_i,\widetilde{\mA}^{(k)}_i\Big)\bigg)},
\end{equation}
where $\widetilde{\mA}^{(t)}_i \in \mathbb{R}^{B\times B}$ is the temporal attention of node $i$ on $G^{t}$. To compute Equation~\ref{eq:continuity_reg} for all of the nodes instead of one-by-one, we form a block diagonal matrix with each diagonal block be the attention corresponding to each node, say $\widetilde{\mA}^{(t)}_{\text{block}}=\text{diag}\Big(\widetilde{\mA}^{(t)}_0,...,\widetilde{\mA}^{(t)}_N\Big)$.  
\subsection{Buffer-based live-update}
After we obtain the node embedding $\mH^{'(t)}\in\sR^{N\times K}$ for the current snapshot, DyExplainer uses an MLP to predict the probability of a future edge from node $i$ to $j$. We compute a cross entropy loss $\gL_{ce}$ between the predictions and the edge labels at the future snapshot. After all, we have the objective function
\begin{equation}
    \label{eq:obj}
    \gL = (1-\alpha-\beta)\gL_{ce}+\alpha  \mathcal{L}_{cons}+ \beta \mathcal{L}_{cont}.
\end{equation}
Inspired by ROLAND~\cite{you2022roland}, we develop a buffer-based live-update algorithm to train the model. The key idea is to balance the efficiency and the aggregation from historical embeddings. Specifically, we update the backbone lively and fine-tuning the attention modules with the node embedding from the buffer. Note that the backbone is updated based on cross-entropy loss because without attention module, the terms $\gL_{cons}$ and $\gL_{cont}$ are zeros. We provide the details in Algorithm \ref{alg:graph}.
\begin{algorithm}
    \centering
    \caption{Buffer-based live-update algorithm.}\label{alg:graph}
\begin{small}
    \begin{algorithmic}[1]
        \STATE {\bfseries Input:} Dynamic graphs $\gG=(G^{(1)},...,G^{(T)})$, link prediction labels $y_1,...y_T$, number of snapshots $T$, maximum fine-tuning epoch $E$, maximum buffer size $B$, trade-off parameters $\alpha$ and $\beta$, the DyExplainer model.
        \STATE Initialize the node state $H^{(0)}$ and an empty queue as a buffer with size 0;
        \FOR{$t=1,...,T-1$}
            \STATE Train and update the backbone module based on $y_{t}$ with early stopping and get $H^{(t)}$;
            \IF{Buffer size < B}
                \STATE Insert $\mH^{(t)}$ into the buffer;
            \ELSE
                \STATE Delete $\mH^{(t-B)}$ and insert $\mH^{(t)}$;
            \ENDIF
            \FOR{$e = 1,...,E$}
                \STATE Input the node embedding stored in the buffer to the structural attention and get $\widetilde{\mH}^{(t)}$, as described in Section~\ref{sec:str_agg};
                \STATE Compute $\gL_{cons}$ in equation~\ref{eq:consistency_reg};
                \STATE Input the structural node embedding to the temporal attention and get $\mH^{'(t)}$, as described in Section~\ref{sec:temp_agg};
                \STATE Compute $\gL_{cont}$ in equation~\ref{eq:continuity_reg};
                \STATE Compute $\gL$ in equation~\ref{eq:obj};
                \STATE Train and Update the attention modules based on $y_{t}$. 
            \ENDFOR
    \ENDFOR
\end{algorithmic}
\end{small}
\end{algorithm}

\normalsize
\section{Experiments}
\label{sec:exp}
To evaluate the performance of the DyExplainer, we compare it against state-of-the-art baselines. Our findings demonstrate the effectiveness in model generalization for link prediction
Furthermore, we quantitatively validate the accuracy of its explanations. We also delve deeper with ablation studies and case studies, offering a deeper understanding of the proposed method.

\begin{table*}[t]
\setlength\tabcolsep{3pt}
\centering
\caption{\textbf{Comparison of MRR 
for the DyExplainer and the baselines. Standard deviations are obtained by
repeating each model training 5 times. For each data set, the two best cases are boldface.}}
\begin{footnotesize}
\vspace{-1mm}
\scalebox{1.25}{
    \begin{tabular}{ccccccc}
    \toprule
    &\multicolumn{1}{c}{AS-733}&\multicolumn{1}{c}{Reddit-Title}&\multicolumn{1}{c}{Reddit-Body}&\multicolumn{1}{c}{UCI-Message}&\multicolumn{1}{c}{Bitcoin-OTC}&\multicolumn{1}{c}{Bitcoin-Alpha}\\ 
    \midrule
    EvolveGCN-H&0.263 $\pm$ 0.098&0.156 $\pm$ 0.121&0.072 $\pm$ 0.010&0.055 $\pm$ 0.011&0.081 $\pm$ 0.025&0.054 $\pm$ 0.019\\
    EvolveGCN-O&0.180 $\pm$ 0.104 &0.015 $\pm$ 0.019&0.093 $\pm$ 0.022&0.028 $\pm$ 0.005&0.018 $\pm$ 0.008&0.005 $\pm$ 0.006\\
    GCRN-GRU&\textbf{0.337 $\pm$ 0.001}&0.328 $\pm$ 0.005&0.204 $\pm$ 0.005&0.095 $\pm$ 0.013&0.163 $\pm$ 0.005&0.143 $\pm$ 0.004\\
    GCRN-LSTM&0.335 $\pm$ 0.001&0.343 $\pm$ 0.006&0.209 $\pm$ 0.003&\textbf{0.107 $\pm$ 0.004}&0.172 $\pm$ 0.013& 0.146 $\pm$ 0.008\\
    GCRN-Baseline&0.321 $\pm$ 0.002&0.342 $\pm$ 0.004&0.202 $\pm$ 0.002&0.090 $\pm$ 0.011& 0.176 $\pm$ 0.005& \textbf{0.152} $\pm$ \textbf{0.005}\\
    TGCN&0.335 $\pm$ 0.001&0.382 $\pm$ 0.005& 0.234 $\pm$ 0.004 &0.080 $\pm$ 0.015&0.080 $\pm$ 0.006&0.060 $\pm$ 0.014\\ 
    ROLAND& 0.330 $\pm$ 0.004& \textbf{0.384 $\pm$ 0.013} & \textbf{0.342 $\pm$ 0.008} & 0.090 $\pm$ 0.010 & \textbf{0.189 $\pm$ 0.008} & 0.147 $\pm$ 0.006\\ 
    \midrule
    DyExplainer & \textbf{0.341} $\pm$ \textbf{0.000} & \textbf{0.383 $\pm$ 0.002} & \textbf{0.335 $\pm$ 0.010}  & \textbf{0.109 $\pm$ 0.004} & \textbf{0.194 $\pm$   0.002}& \textbf{0.164} $\pm$ \textbf{0.002} \\
    \bottomrule
    \end{tabular}
    }
\label{tab:linkpred}
\end{footnotesize}
\end{table*}

\subsection{Experimental Setup}
\textbf{Datasets.} The experiments are performed on six widely used data sets in the following list. The data set (1) AS-733 is an autonomous systems dataset of traffic flows among routers comprising the Internet~\cite{leskovec2005graphs}. (2) Reddit-Title and (3) Reddit-Body are networks of subreddit-to-subreddit hyperlinks extracted from posts. The posts contain hyperlinks that connect one subreddit to another. The edge label shows if the source post expresses negativity towards the target post~\cite{kumar2018community}. (4) UCI-Message is composed of private communications exchanged on an online social network system among students~\cite{panzarasa2009patterns}.
The data sets Bitcoin-OTC and Bitcoin-Alpha consist of "who-trusts-whom" networks of individuals who engage in trading on these platforms~\cite{kumar2018rev2, kumar2016edge}. 

\noindent\textbf{Baselines.} We compare DyExplainer with both 
link prediction methods
and explainable methods. For link prediction methods, we use 7 state-of-the-art dynamic GNNs.
The (1) EvolveGCN-H and (2) EvolveGCN-O models employ an RNN to dynamically adapt the weights of internal GNNs, enabling the GNN to change during testing~\cite{pareja2020evolvegcn}. (3) T-GCN integrates a GNN into the GRU cell by replacing the linear transformations in GRU with graph convolution operators~\cite{zhao2019t}. The (4) GCRN-GRU and (5) GCRN-LSTM methods are widely adopted baselines that are generalized to capture temporal information by incorporating either a GRU or an LSTM layer. GCRN uses a ChebNet~\cite{defferrard2016convolutional} for spatial information and separate GNNs to compute different gates of RNNs. The (6) GCRN-Baseline first builds node features using a Chebyshev spectral graph convolution layer to capture spatial information, then feeds these features into an LSTM cell to extract temporal information~\cite{seo2018structured}. The (7) ROLAND views the node embeddings at different layers in the GNNs as hierarchical node states, which it updates recurrently over time. It integrates advanced design features from static GNNs and enables lively updating. Throughout our experiments, we use the GRU-based ROLAND which is shown to perform better than the others~\cite{you2022roland}. 

To evaluate the effectiveness of 
explainability, we compare DyExplainer with GNNExplainer~\cite{ying2019gnnexplainer} and a gradient-based
method (Grad). (1) GNNExplainer is a post-hoc state-of-the-art method providing explanations for every single instance. (2). Grad learns weights of edges by computing gradients of the model’s objective function w.r.t. the adjacency matrix.

\noindent\textbf{Metrics.} For measuring link prediction performance, we use the standard mean reciprocal rank (MRR). For
evaluating the faithfulness of explainability, we mainly use the Fidelity score of probability~\cite{yuan2021explainability}. We let $\gG^t=\{\gV_t, \gE_t\}$ be the graph at the snapshot $t$, with $\gV_t$ a set of vertices and $\gE_t$ a set of edges. For snapshot $t$, there is a set of edges $\gE'_t$ needed to make predictions. 
After training the DyExplainer, we obtain an explanation mask $\mM_t \in \{0, 1\}^{n\times n}$ for the snapshot $\gG^t$, with each element 0 or 1 to indicate whether the corresponding edge is identified as important. According to $\mM_t$ we obtain the important edges in $\gG^t$ to create a new graph $\hat{\gG}^t$. The Fidelity score is computed as
\begin{small}
\begin{equation}
\label{eq:fidelity}
    Fidelity = \frac{1}{\left|\gE'_{t}\right|}\sum_{i=1}^{\left|\gE'_{t}\right|} \left|f\big(\gG^t \big)_{e'_i}- f\big(\hat{\gG}^t \big)_{e'_i}\right|,
\end{equation}
\end{small}
 \noindent  where $\left|\gE'_{t}\right|$ is the number of edges need to predict, $f\big(\gG^t\big)_{e'_i}$ means the predicted probability of edge $e'_i \in \gE'_{t}$. We follow~\cite{pope2019explainability,yuan2022explainability} to compute the Fidelity scores at different sparsity given by
 \begin{small}
\begin{equation}
\label{eq:sparsity}
    Sparsity = 1-\frac{\left|\mM_t\right|}{\left|\gE_t\right|},
\end{equation}
\end{small}
where $\left|\mM_t\right|$ is the number of important edges identified in $\mM_t$ and $\left|\gE_t\right|$ means the number of edges in $\gG_t$.

\noindent\textbf{Implementation details.}
To ensure a fair comparison, all methods were trained through live updating as outlined in~\cite{you2022roland}. The evaluation is supposed to happen at each snapshot, however, models training on streaming data with early-stopping usually do not converge in the early epochs. Therefore, we report the average of MRR of the most recent 60\% snapshots. The hyper-parameter space of the DyExplainer model is similar to the hyper-parameter space of the underlying static GNN layers. For all methods, the node state hidden dimensions are set to 128 with GNN layers featuring skip connection, sum aggregation, and batch normalization. The DyExplainer is trained for a maximum of 100 epochs in each time step before early stopping, and its explainable module is fine-tuned for several additional epochs. For the attention modules, the slope in $\text{LeakyRelU}$ is set to $0.2$. We follow the practice in~\cite{jang2016categorical} to adopt the exponential decay strategy by starting the training with a high temperature of 1.0, and annealing to a small value of 0.1. For each dataset, we follow~\cite{you2022roland} of the parameter settings in the backbone. we use grid search to tune the hyperparameters of DyExplainer. Specifically, we tune: (1) The hidden dimension for structural and temporal attention modules (8, 16); (2) the learning rate for live-update the backbone (0.001 to 0.02); (3) the learning rate for fine-tuning the explainable module (0.001 to 0.1); (4) buffer size of the explainable module (3 to 20); (5) trade-off parameters $\alpha$ for temporal regularization and $\beta$ for structural regularization (0 to 1).

\noindent\textbf{Computing environment.} We implemented all models using PyTorch~\cite{paszke2019pytorch}, PyTorch Geometric~\cite{fey2019fast}, and Scikit-learn~\cite{pedregosa2011scikit}. All data sets used in the experiments are obtained from~\cite{snapnets}. We conduct the experiments on a server with four NVIDIA RTX A6000 GPUs (48GB memory each).

\subsection{Link Prediction Performance}
Table~\ref{tab:linkpred} showcases the MRR results for all compared methods on all datasets, which were all trained through live updating. The results are obtained by averaging the MRR of the most recent 60\% snapshots on the test datasets. The buffer size for DyExplainer was set to 5 across all datasets, and the attention module was fine-tuned for 4 epochs. The ROLAND method outperforms the other baselines on most datasets. DyExplainer surpasses the best baseline on 4 datasets, showing a 7.89\% improvement on Bitcoin-Alpha, and performs similarly to the best baseline on the remaining datasets. This effectively demonstrates the Rashomon set theory~\cite{ranshomon1,ranshomon2}, and highlights the benefits of seeking a simple and understandable model, which can result in improved robustness and better overall performance.

\subsection{Ablation Study}
\label{sec:ablation}
\begin{figure}
 \centering
 \includegraphics[width = 0.35\textwidth]{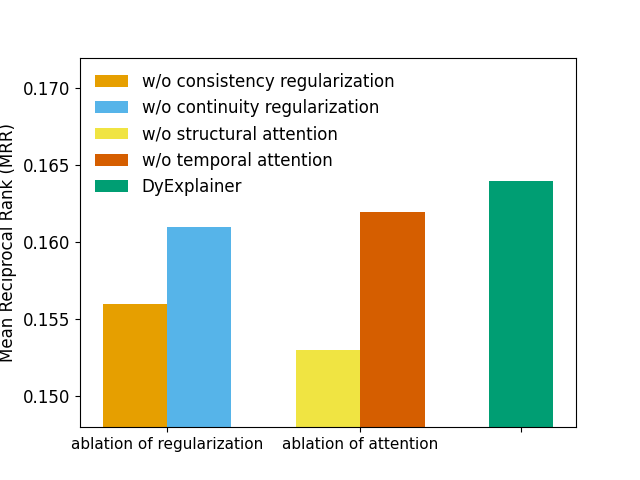}
 \caption{Ablation studies on Bitcoin-Alpha.}
 \vskip -5pt
 \label{fig:ablation}
 \vspace{-3mm}
\end{figure}

To present deep insights into the proposed method, we conducted multiple ablation studies on the Bitcoin-Alpha dataset to empirically verify the effectiveness of the proposed explainable aggregations and the contrastive regularizations. Specifically, we
compare DyExplainer with variants. (1) w/o structural attention and w/o temporal attention are the DyExplainer without one of the attention modules in the explainable aggregations. (2) w/o consistency regularization and w/o continuity regularization refers to DyExplainer without one of the contrastive regularization terms in the loss. 
Results are shown in Figure~\ref{fig:ablation}.

From the figure, we observe that: (1) the performance of w/o structural attention is much worse than w/o temporal attention; (2)  the performance of w/o consistency regularization is much worse than w/o continuity regularization. It is evident from both (1) and (2) that the topological information within a single snapshot holds greater sway over the prediction outcome, as compared to the temporal dependencies that exist between snapshots. (3) The results of the ablation studies performed on regularization and attention further reinforce the superiority of our approach over these alternatives.

\subsection{Explanations Performance}
In order to demonstrate the reliability of the explanations provided by DyExplainer, we conduct quantitative evaluations that compare our approach with various baselines across three datasets characterized by a limited number of edges: UCI-Message, Bitcoin-OTC, and Bitcoin-Alpha. 
Specifically, we adopt the Fidelity vs. Sparsity metrics for our evaluation, in accordance with the methodology described in~\cite{pope2019explainability,yuan2021explainability}. The Fidelity metric assesses the accuracy with which the explanations reflect the significance of various factors to the model's predictions, while the Sparsity metric quantifies the proportion of structures that are deemed critical by the explanation techniques.  For a fair comparison, for all three methods, we use a trained dynamic GNN~\cite{you2022roland} as the base model to calculate the predicted probability.
\begin{figure}[t]
 \centering
 \includegraphics[width = 0.48\textwidth]{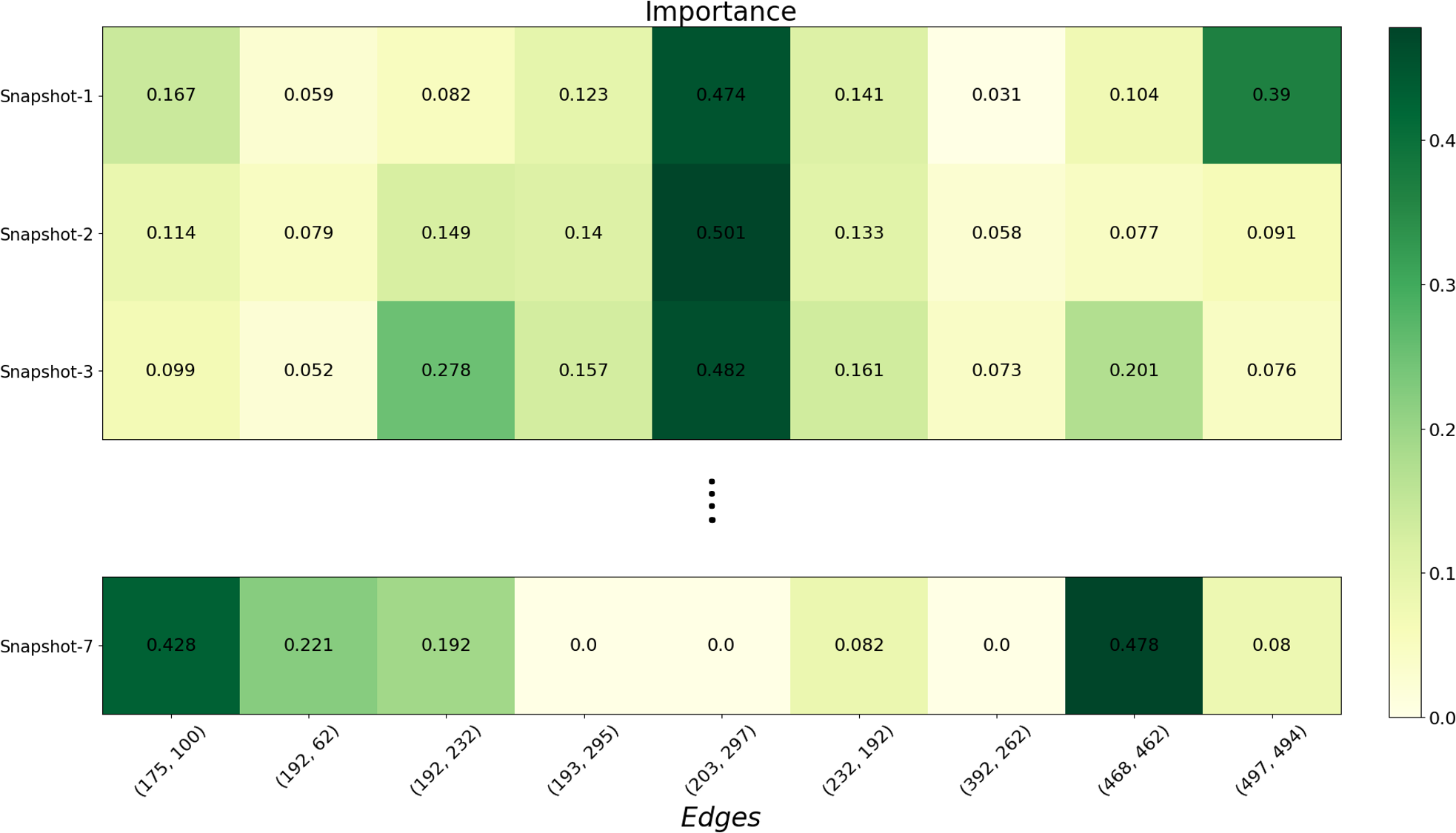}
 \vspace{-2em}
 \caption{Heat map of edge importance over time. We sample some edges shared by continuous snapshots and show their attention values. 0.0 means this edge does not appear in this snapshot.}
 \vspace{-2em}
 \label{fig:case_study}
\end{figure}
\begin{figure*}[t]
\centering\subfigure{\includegraphics[width=.28\linewidth]{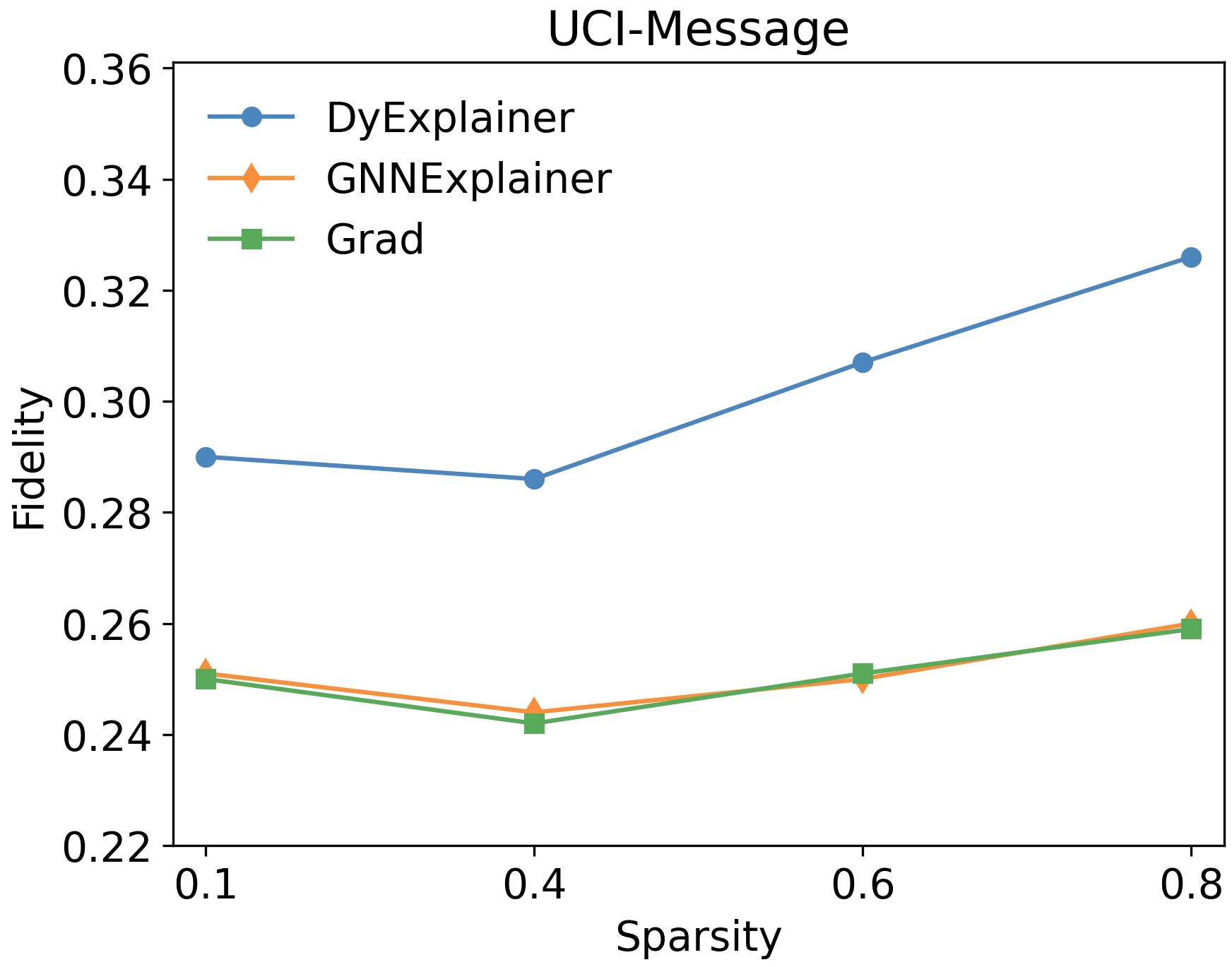}}
    \subfigure{ \includegraphics[width=.28\linewidth]{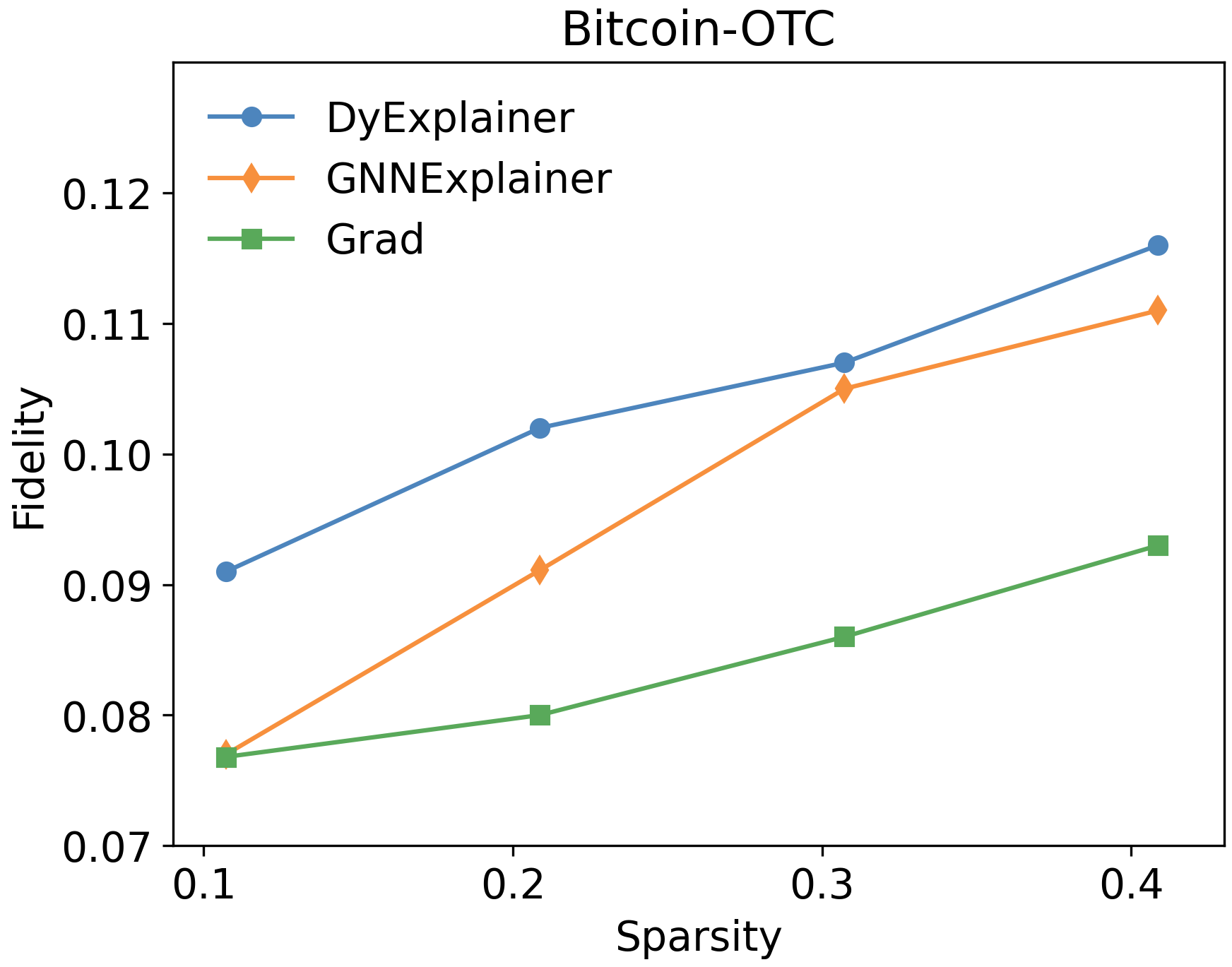}}
\subfigure{\includegraphics[width=.28\linewidth]{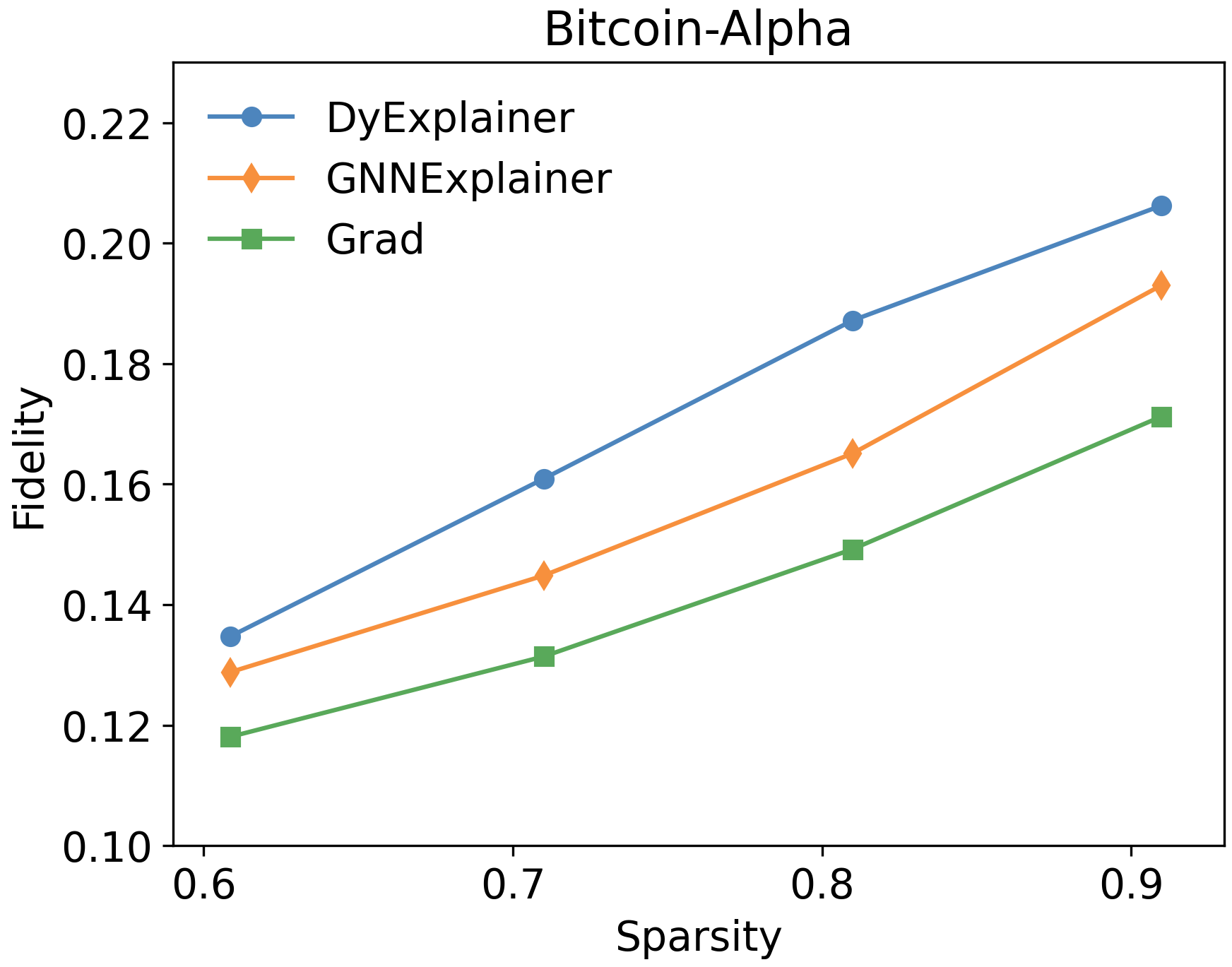}}
    \caption{The quantitative studies for different explanation methods on UCI-Message, Bitcoin-OTC, and Bitcoin-Alpha.}
    \label{fig:fidelity}
    \vspace{-3mm}
  \end{figure*}

The baseline method, GNNExplainer, was not originally developed for dynamic graph settings. Nevertheless, for a fair comparison, we assess the Fidelity of both DyExplainer and GNNExplainer using only the graph at the final snapshot, despite the fact that DyExplainer provides explanations for all snapshots stored in a buffer. GNNExplainer identifies the most influential nodes within a k-hop neighborhood and generates a mask to highlight these nodes for a given prediction node. As our DyExplainer provides a global attention mechanism for the entire graph, for a fair comparison, we calculate the average Fidelity of each mask produced by GNNExplainer with respect to all of the edges. To obtain the Fidelity and Sparsity metrics for each mask, we select a subset of the top-ranked edges based on the weights from the explanation mask and use this subset to create a new graph. The Fidelity of an edge is defined as the difference between the predicted probability of that edge on the new graph and the global graph. For DyExplainer, we directly select the top-ranked edges based on the attention weights to form a new graph, as there is only one global attention mechanism that provides the explanation.

The comparison of Fidelity and Sparsity is presented in Figure~\ref{fig:fidelity}. The evaluation of various methods is based on their Fidelity scores under comparable levels of Sparsity, as the Sparsity level cannot be precisely controlled. From the figure, we see that for all three datasets, the Fidelity of these methods increases as Sparsity increases. This is due to the calculation of Fidelity as the difference between the predicted probability of the model on the reduced graph and the original graph, leading to higher Fidelity values with greater Sparsity. Furthermore, GNNExplainer demonstrates superior Fidelity compared to Grad on Bitcoin-OTC and Bitcoin-Alpha, while performing similarly on UCI-Message. Additionally, DyExplainer outperforms both methods on all three datasets at varying Sparsity levels, revealing that it provides more accurate explanations.

\subsection{Case Studies}

To show the evolving and continuity of underlying patterns the DyExplainer detects from the dynamic graph,
we visualize the attention values of some edges at different snapshots on the Bitcoin-Alpha data set in Figure~\ref{fig:case_study}. From this figure, we observe that the edges $(192, 62)$, $(193, 295)$, $(203, 297)$ and $(232, 192)$ has close importance on snapshots 1-3. It indicates the temporal continuity of edge importance exists in dynamic graphs and our intuition is reasonable. The edges $(175, 100)$, $(192, 62)$, and $(468, 462)$ are less important at snapshots 1-3 while becoming more important at snapshot 7. It infers that the importance of an edge is evolving along with the time steps.

\begin{figure}[htbp]
 \centering
 \includegraphics[width = 0.45\textwidth]{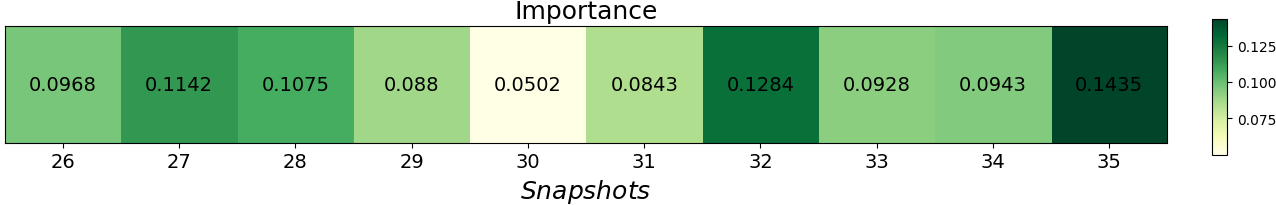}
 \vspace{-1mm}
 \caption{Heat map of temporal importance over time. The current snapshot is at 35. The buffer size is set to 10. It means we consider the importance of the previous snapshot from 27 to 34.}
 \vspace{-2mm}
\label{fig:case_study2}
\end{figure}

\begin{figure}[htbp]
 \centering
\includegraphics[width = 0.35\textwidth]{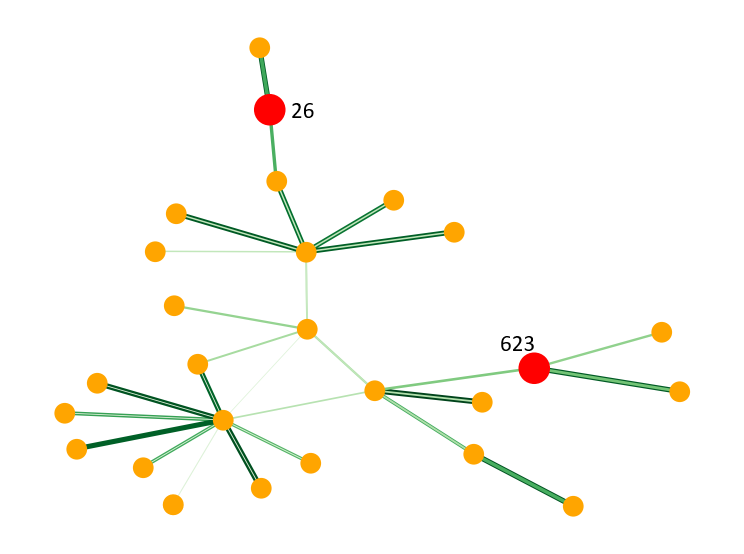}
\vspace{-2mm}
\caption{Local structure of node 26 and 623 (red nodes) on snapshot 32.  Orange points are their 3-hop neighborhoods. The green edges are obtained from the structural attention weights at snapshot 32.}
\vspace{-3mm}
\label{fig:case_study3}
\end{figure}

We further visualize the temporal attention to show which historical snapshot has more influence on the current. The importance of snapshots provided by the temporal attention is shown in Figure~\ref{fig:case_study2}. From the figure, we see that the most important snapshot to the current is snapshot 32.  
To see how the patterns have an effect, we randomly pick an edge $(623, 26)$ at snapshot 35 to investigate the local structure of this edge at snapshot 32. The visualization of nodes 623 and 26 are shown in Figure~\ref{fig:case_study3}. Note that these two nodes are unconnected at snapshot 32. The widths of edges are according to the weights of structural attention. Thicker ones mean large weights. From the figure, we can find that there are three clusters of substructures with larger weights, which can be viewed as important patterns that DyExplainer detects.

\vspace{-1mm}
\section{Related Work}
\label{related:related}
The goal of explainability in GNNs is to provide transparency and accountability in the predictions of the models, especially when they are used in critical applications such as detecting the fraud~\cite{liu2021pick} or medical diagnosis~\cite{liu2020heterogeneous}. Recently, many works are proposed to explain the GNN predictions, with a focus on diverse facets of the models from various perspectives. According to the type of explanation they provide, the methodologies can be categorized into two main classes: instance-level and model-level methods~\cite{yuan2022explainability}. 

The instance-level methods explain the GNN models by identifying the most influential features of the graph to the given prediction. Among them, some methods employ the gradients or the feature values
to indicate the importance of features. SA~\cite{baldassarre2019explainability} computes the gradient value as
the importance score for each input feature. While it is easy to compute by the back-propagation, the results 
cannot accurately capture the importance of each feature due to the output changing minimally with respect to the input change, and a gradient value is hard to reflect the input contribution. CAM~\cite{pope2019explainability} maps the node features in the final layer to the input space to identify important nodes. However, the representation from the final layer of GNN may not reflect the node contribution because the feature distribution may change after mapping by a neural network. Another kind of method aims to provide instance-level explanations by a surrogate model that approximates the
predictions of the original GNN model. GraphLime~\cite{huang2022graphlime} provides a model-agnostic local explanation framework for the node classification task. It adopts the node feature and predicted labels in the K-hop neighbors of the predicted node and trains an HSIC Lasso. 
GNNExplainer~\cite{ying2019gnnexplainer} takes a trained
GNN and its predictions as inputs to provide explanations for a given instance, e.g. a
node or a graph. The explanation includes a compact subgraph structure and a small subset of node
features that are crucial in GNN’s prediction for the target instance.
However, the explanation provided by GNNExplainer is limited to a single instance, making GNNExplainer difficult to be applied in the inductive setting because the explanations are hard
to generalize to other unexplained nodes. PGExplainer~\cite{luo2020parameterized} is proposed to provide a global understanding of predictions made by GNNs. It models the underlying
structure as edge distributions where the explanatory graph is sampled. To explain the predictions of multiple instances, the generation process in PGExplainer is parameterized by a
neural network. Similar to the
PGExplainer, GraphMask~\cite{schlichtkrull2020interpreting} trains a classifier to predict whether an edge
can be dropped without affecting the original predictions. As a post-hoc method, GraphMask obtains an edge mask for each GNN layer. SubgraphX~\cite{yuan2021explainability} explains the GNN predictions as an efficient exploration of different subgraphs with Monte Carlo tree search. It adopts Shapley values
as a measure of subgraph importance that also capture the interactions among different subgraphs. All of these methods are to provide explanations with respect to the GNN predictions. Different from them, this work provides a high-level understanding to explain the GNN models. The model-level methods that study what graph patterns can lead to a certain GNN behavior, e.g., the improvement of performance, is particularly related to us. There are fewer studies in this field. The method XGNN~\cite{yuan2020xgnn} aims to explain GNNs by training a graph generator so that
the generated graph patterns maximize a certain prediction of the
model. 
The explanations by XGNN are general that provide a global understanding of the trained GNNs.

Current explainers for GNNs are limited to static graphs, hindering their application in dynamic scenarios. The explanation of dynamic GNNs is an under-studied area. Indeed, there is a recent work~\cite{xie2022explaining} attempted to provide explanations on dynamic graphs by exploring backward relevance, but it is limited to a specific model TGCN. Besides, the work in~\cite{he2022explainer} considers explaining time series predictions by temporal GNNs. 
The work~\cite{fan2021gcn} learns attention weights for
a linear combination of node representations dynamic graphs, but it explains the model by detecting the node importance. There are distinct challenges in explaining dynamic GNNs in general. Firstly, existing explanation methods mainly focus on identifying the important parts of the data in relation to the GNN predictions. However, dynamic GNNs make predictions for each snapshot, making it difficult 
to provide explanations for a particular prediction as it depends on all previous snapshots. Secondly, it is challenging to provide a generalizable solution to interpret various types of dynamic GNNs. DyExplainer, on the other hand, provides model-level explanations for all dynamic GNNs, overcoming these limitations.

\vspace{-3mm}
\section{Conclusion}
\vspace{-1mm}
\label{sec:con}


We present DyExplainer, a pioneering approach to explaining dynamic GNNs in real-time. DyExplainer leverages a dynamic GNN backbone to extract representations at each snapshot, concurrently exploring structural relationships and temporal dependencies through a sparse attention mechanism. To ensure structural consistency and temporal continuity in the explanation, our approach incorporates contrastive learning techniques and a buffer-based live-updating scheme. The results of our experiments showcase the superiority of DyExplainer, providing a faithful explanation of the model predictions while concurrently improving the accuracy of the model, as evidenced by the link prediction task.

\bibliographystyle{ACM-Reference-Format}
\bibliography{sample-sigconf.bbl}


\begin{thebibliography}{52}


\ifx \showCODEN    \undefined \def \showCODEN     #1{\unskip}     \fi
\ifx \showDOI      \undefined \def \showDOI       #1{#1}\fi
\ifx \showISBNx    \undefined \def \showISBNx     #1{\unskip}     \fi
\ifx \showISBNxiii \undefined \def \showISBNxiii  #1{\unskip}     \fi
\ifx \showISSN     \undefined \def \showISSN      #1{\unskip}     \fi
\ifx \showLCCN     \undefined \def \showLCCN      #1{\unskip}     \fi
\ifx \shownote     \undefined \def \shownote      #1{#1}          \fi
\ifx \showarticletitle \undefined \def \showarticletitle #1{#1}   \fi
\ifx \showURL      \undefined \def \showURL       {\relax}        \fi
\providecommand\bibfield[2]{#2}
\providecommand\bibinfo[2]{#2}
\providecommand\natexlab[1]{#1}
\providecommand\showeprint[2][]{arXiv:#2}

\bibitem[Baldassarre and Azizpour(2019)]%
        {baldassarre2019explainability}
\bibfield{author}{\bibinfo{person}{Federico Baldassarre} {and}
  \bibinfo{person}{Hossein Azizpour}.} \bibinfo{year}{2019}\natexlab{}.
\newblock \showarticletitle{Explainability techniques for graph convolutional
  networks}.
\newblock \bibinfo{journal}{\emph{Preprint arXiv:1905.13686}}
  (\bibinfo{year}{2019}).
\newblock


\bibitem[Cho et~al\mbox{.}(2014)]%
        {cho2014properties}
\bibfield{author}{\bibinfo{person}{Kyunghyun Cho}, \bibinfo{person}{Bart
  Van~Merri{\"e}nboer}, \bibinfo{person}{Dzmitry Bahdanau}, {and}
  \bibinfo{person}{Yoshua Bengio}.} \bibinfo{year}{2014}\natexlab{}.
\newblock \showarticletitle{On the properties of neural machine translation:
  Encoder-decoder approaches}.
\newblock \bibinfo{journal}{\emph{arXiv preprint arXiv:1409.1259}}
  (\bibinfo{year}{2014}).
\newblock


\bibitem[Defferrard et~al\mbox{.}(2016)]%
        {defferrard2016convolutional}
\bibfield{author}{\bibinfo{person}{Micha{\"e}l Defferrard},
  \bibinfo{person}{Xavier Bresson}, {and} \bibinfo{person}{Pierre
  Vandergheynst}.} \bibinfo{year}{2016}\natexlab{}.
\newblock \showarticletitle{Convolutional neural networks on graphs with fast
  localized spectral filtering}.
\newblock \bibinfo{journal}{\emph{NIPS}}  \bibinfo{volume}{29}
  (\bibinfo{year}{2016}).
\newblock


\bibitem[Fan et~al\mbox{.}(2021)]%
        {fan2021gcn}
\bibfield{author}{\bibinfo{person}{Yucai Fan}, \bibinfo{person}{Yuhang Yao},
  {and} \bibinfo{person}{Carlee Joe-Wong}.} \bibinfo{year}{2021}\natexlab{}.
\newblock \showarticletitle{Gcn-se: Attention as explainability for node
  classification in dynamic graphs}. In \bibinfo{booktitle}{\emph{2021 IEEE
  International Conference on Data Mining (ICDM)}}. IEEE,
  \bibinfo{pages}{1060--1065}.
\newblock


\bibitem[Fey and Lenssen(2019)]%
        {fey2019fast}
\bibfield{author}{\bibinfo{person}{Matthias Fey} {and}
  \bibinfo{person}{Jan~Eric Lenssen}.} \bibinfo{year}{2019}\natexlab{}.
\newblock \showarticletitle{Fast graph representation learning with PyTorch
  Geometric}.
\newblock \bibinfo{journal}{\emph{RLGM@ICLR}} (\bibinfo{year}{2019}).
\newblock


\bibitem[He et~al\mbox{.}(2022)]%
        {he2022explainer}
\bibfield{author}{\bibinfo{person}{Wenchong He}, \bibinfo{person}{Minh~N Vu},
  \bibinfo{person}{Zhe Jiang}, {and} \bibinfo{person}{My~T Thai}.}
  \bibinfo{year}{2022}\natexlab{}.
\newblock \showarticletitle{An explainer for temporal graph neural networks}.
  In \bibinfo{booktitle}{\emph{GLOBECOM}}. IEEE, \bibinfo{pages}{6384--6389}.
\newblock


\bibitem[Hochreiter and Schmidhuber(1997)]%
        {hochreiter1997long}
\bibfield{author}{\bibinfo{person}{Sepp Hochreiter} {and}
  \bibinfo{person}{J{\"u}rgen Schmidhuber}.} \bibinfo{year}{1997}\natexlab{}.
\newblock \showarticletitle{Long short-term memory}.
\newblock \bibinfo{journal}{\emph{Neural computation}} \bibinfo{volume}{9},
  \bibinfo{number}{8} (\bibinfo{year}{1997}), \bibinfo{pages}{1735--1780}.
\newblock


\bibitem[Huang et~al\mbox{.}(2022)]%
        {huang2022graphlime}
\bibfield{author}{\bibinfo{person}{Qiang Huang}, \bibinfo{person}{Makoto
  Yamada}, \bibinfo{person}{Yuan Tian}, \bibinfo{person}{Dinesh Singh}, {and}
  \bibinfo{person}{Yi Chang}.} \bibinfo{year}{2022}\natexlab{}.
\newblock \showarticletitle{Graphlime: Local interpretable model explanations
  for graph neural networks}.
\newblock \bibinfo{journal}{\emph{IEEE Transactions on Knowledge and Data
  Engineering}} (\bibinfo{year}{2022}).
\newblock


\bibitem[Jang et~al\mbox{.}(2016)]%
        {jang2016categorical}
\bibfield{author}{\bibinfo{person}{Eric Jang}, \bibinfo{person}{Shixiang Gu},
  {and} \bibinfo{person}{Ben Poole}.} \bibinfo{year}{2016}\natexlab{}.
\newblock \showarticletitle{Categorical reparameterization with
  gumbel-softmax}.
\newblock \bibinfo{journal}{\emph{Preprint arXiv:1611.01144}}
  (\bibinfo{year}{2016}).
\newblock


\bibitem[Kipf and Welling(2016)]%
        {kipf2016semi}
\bibfield{author}{\bibinfo{person}{Thomas~N Kipf} {and} \bibinfo{person}{Max
  Welling}.} \bibinfo{year}{2016}\natexlab{}.
\newblock \showarticletitle{Semi-supervised classification with graph
  convolutional networks}.
\newblock \bibinfo{journal}{\emph{Preprint arXiv:1609.02907}}
  (\bibinfo{year}{2016}).
\newblock


\bibitem[Kumar et~al\mbox{.}(2018a)]%
        {kumar2018community}
\bibfield{author}{\bibinfo{person}{Srijan Kumar}, \bibinfo{person}{William~L
  Hamilton}, \bibinfo{person}{Jure Leskovec}, {and} \bibinfo{person}{Dan
  Jurafsky}.} \bibinfo{year}{2018}\natexlab{a}.
\newblock \showarticletitle{Community interaction and conflict on the web}. In
  \bibinfo{booktitle}{\emph{WWW}}. \bibinfo{pages}{933--943}.
\newblock


\bibitem[Kumar et~al\mbox{.}(2018b)]%
        {kumar2018rev2}
\bibfield{author}{\bibinfo{person}{Srijan Kumar}, \bibinfo{person}{Bryan Hooi},
  \bibinfo{person}{Disha Makhija}, \bibinfo{person}{Mohit Kumar},
  \bibinfo{person}{Christos Faloutsos}, {and} \bibinfo{person}{VS
  Subrahmanian}.} \bibinfo{year}{2018}\natexlab{b}.
\newblock \showarticletitle{Rev2: Fraudulent user prediction in rating
  platforms}. In \bibinfo{booktitle}{\emph{WSDM}}. \bibinfo{pages}{333--341}.
\newblock


\bibitem[Kumar et~al\mbox{.}(2016)]%
        {kumar2016edge}
\bibfield{author}{\bibinfo{person}{Srijan Kumar}, \bibinfo{person}{Francesca
  Spezzano}, \bibinfo{person}{VS Subrahmanian}, {and} \bibinfo{person}{Christos
  Faloutsos}.} \bibinfo{year}{2016}\natexlab{}.
\newblock \showarticletitle{Edge weight prediction in weighted signed
  networks}. In \bibinfo{booktitle}{\emph{ICDM}}. IEEE,
  \bibinfo{pages}{221--230}.
\newblock


\bibitem[Leskovec et~al\mbox{.}(2005)]%
        {leskovec2005graphs}
\bibfield{author}{\bibinfo{person}{Jure Leskovec}, \bibinfo{person}{Jon
  Kleinberg}, {and} \bibinfo{person}{Christos Faloutsos}.}
  \bibinfo{year}{2005}\natexlab{}.
\newblock \showarticletitle{Graphs over time: densification laws, shrinking
  diameters and possible explanations}. In \bibinfo{booktitle}{\emph{ACM
  SIGKDD}}. \bibinfo{pages}{177--187}.
\newblock


\bibitem[Leskovec and Krevl(2014)]%
        {snapnets}
\bibfield{author}{\bibinfo{person}{Jure Leskovec} {and} \bibinfo{person}{Andrej
  Krevl}.} \bibinfo{year}{2014}\natexlab{}.
\newblock \bibinfo{title}{{SNAP Datasets}: {Stanford} Large Network Dataset
  Collection}.
\newblock \bibinfo{howpublished}{\url{http://snap.stanford.edu/data}}.
\newblock


\bibitem[Liang et~al\mbox{.}(2022)]%
        {Liangke_Survey}
\bibfield{author}{\bibinfo{person}{Ke Liang}, \bibinfo{person}{Lingyuan Meng},
  \bibinfo{person}{Meng Liu}, \bibinfo{person}{Yue Liu},
  \bibinfo{person}{Wenxuan Tu}, \bibinfo{person}{Siwei Wang},
  \bibinfo{person}{Sihang Zhou}, \bibinfo{person}{Xinwang Liu}, {and}
  \bibinfo{person}{Fuchun Sun}.} \bibinfo{year}{2022}\natexlab{}.
\newblock \showarticletitle{Reasoning over different types of knowledge graphs:
  Static, temporal and multi-modal}.
\newblock \bibinfo{journal}{\emph{Preprint arXiv:2212.05767}}
  (\bibinfo{year}{2022}).
\newblock


\bibitem[Liu et~al\mbox{.}(2023)]%
        {liu2023deep}
\bibfield{author}{\bibinfo{person}{Meng Liu}, \bibinfo{person}{Yue Liu},
  \bibinfo{person}{Ke Liang}, \bibinfo{person}{Siwei Wang},
  \bibinfo{person}{Sihang Zhou}, {and} \bibinfo{person}{Xinwang Liu}.}
  \bibinfo{year}{2023}\natexlab{}.
\newblock \showarticletitle{Deep Temporal Graph Clustering}.
\newblock \bibinfo{journal}{\emph{Preprint arXiv:2305.10738}}
  (\bibinfo{year}{2023}).
\newblock


\bibitem[Liu et~al\mbox{.}(2021)]%
        {liu2021pick}
\bibfield{author}{\bibinfo{person}{Yang Liu}, \bibinfo{person}{Xiang Ao},
  \bibinfo{person}{Zidi Qin}, \bibinfo{person}{Jianfeng Chi},
  \bibinfo{person}{Jinghua Feng}, \bibinfo{person}{Hao Yang}, {and}
  \bibinfo{person}{Qing He}.} \bibinfo{year}{2021}\natexlab{}.
\newblock \showarticletitle{Pick and choose: a GNN-based imbalanced learning
  approach for fraud detection}. In \bibinfo{booktitle}{\emph{WWW}}.
  \bibinfo{pages}{3168--3177}.
\newblock


\bibitem[Liu et~al\mbox{.}(2020)]%
        {liu2020heterogeneous}
\bibfield{author}{\bibinfo{person}{Zheng Liu}, \bibinfo{person}{Xiaohan Li},
  \bibinfo{person}{Hao Peng}, \bibinfo{person}{Lifang He}, {and}
  \bibinfo{person}{S~Yu Philip}.} \bibinfo{year}{2020}\natexlab{}.
\newblock \showarticletitle{Heterogeneous similarity graph neural network on
  electronic health records}. In \bibinfo{booktitle}{\emph{Big Data}}. IEEE,
  \bibinfo{pages}{1196--1205}.
\newblock


\bibitem[Louizos et~al\mbox{.}(2017)]%
        {louizos2017learning}
\bibfield{author}{\bibinfo{person}{Christos Louizos}, \bibinfo{person}{Max
  Welling}, {and} \bibinfo{person}{Diederik~P Kingma}.}
  \bibinfo{year}{2017}\natexlab{}.
\newblock \showarticletitle{Learning Sparse Neural Networks through $ L\_0 $
  Regularization}.
\newblock \bibinfo{journal}{\emph{Preprint arXiv:1712.01312}}
  (\bibinfo{year}{2017}).
\newblock


\bibitem[Luo et~al\mbox{.}(2020)]%
        {luo2020parameterized}
\bibfield{author}{\bibinfo{person}{Dongsheng Luo}, \bibinfo{person}{Wei Cheng},
  \bibinfo{person}{Dongkuan Xu}, \bibinfo{person}{Wenchao Yu},
  \bibinfo{person}{Bo Zong}, \bibinfo{person}{Haifeng Chen}, {and}
  \bibinfo{person}{Xiang Zhang}.} \bibinfo{year}{2020}\natexlab{}.
\newblock \showarticletitle{Parameterized explainer for graph neural network}.
\newblock \bibinfo{journal}{\emph{Advances in neural information processing
  systems}}  \bibinfo{volume}{33} (\bibinfo{year}{2020}),
  \bibinfo{pages}{19620--19631}.
\newblock


\bibitem[Ma et~al\mbox{.}(2019)]%
        {ma2019graph}
\bibfield{author}{\bibinfo{person}{Yao Ma}, \bibinfo{person}{Suhang Wang},
  \bibinfo{person}{Charu~C Aggarwal}, {and} \bibinfo{person}{Jiliang Tang}.}
  \bibinfo{year}{2019}\natexlab{}.
\newblock \showarticletitle{Graph convolutional networks with eigenpooling}. In
  \bibinfo{booktitle}{\emph{Proceedings of the 25th ACM SIGKDD international
  conference on knowledge discovery \& data mining}}.
  \bibinfo{pages}{723--731}.
\newblock


\bibitem[Maddison et~al\mbox{.}(2016)]%
        {maddison2016concrete}
\bibfield{author}{\bibinfo{person}{Chris~J Maddison}, \bibinfo{person}{Andriy
  Mnih}, {and} \bibinfo{person}{Yee~Whye Teh}.}
  \bibinfo{year}{2016}\natexlab{}.
\newblock \showarticletitle{The concrete distribution: A continuous relaxation
  of discrete random variables}.
\newblock \bibinfo{journal}{\emph{arXiv preprint arXiv:1611.00712}}
  (\bibinfo{year}{2016}).
\newblock


\bibitem[McPherson et~al\mbox{.}(2001)]%
        {mcpherson2001birds}
\bibfield{author}{\bibinfo{person}{Miller McPherson}, \bibinfo{person}{Lynn
  Smith-Lovin}, {and} \bibinfo{person}{James~M Cook}.}
  \bibinfo{year}{2001}\natexlab{}.
\newblock \showarticletitle{Birds of a feather: Homophily in social networks}.
\newblock \bibinfo{journal}{\emph{Annual review of sociology}}
  \bibinfo{volume}{27}, \bibinfo{number}{1} (\bibinfo{year}{2001}),
  \bibinfo{pages}{415--444}.
\newblock


\bibitem[Nauta et~al\mbox{.}(2022)]%
        {nauta2022anecdotal}
\bibfield{author}{\bibinfo{person}{Meike Nauta}, \bibinfo{person}{Jan Trienes},
  \bibinfo{person}{Shreyasi Pathak}, \bibinfo{person}{Elisa Nguyen},
  \bibinfo{person}{Michelle Peters}, \bibinfo{person}{Yasmin Schmitt},
  \bibinfo{person}{J{\"o}rg Schl{\"o}tterer}, \bibinfo{person}{Maurice van
  Keulen}, {and} \bibinfo{person}{Christin Seifert}.}
  \bibinfo{year}{2022}\natexlab{}.
\newblock \showarticletitle{From Anecdotal Evidence to Quantitative Evaluation
  Methods: A Systematic Review on Evaluating Explainable AI}.
\newblock \bibinfo{journal}{\emph{arXiv preprint arXiv:2201.08164}}
  (\bibinfo{year}{2022}).
\newblock


\bibitem[Panzarasa et~al\mbox{.}(2009)]%
        {panzarasa2009patterns}
\bibfield{author}{\bibinfo{person}{Pietro Panzarasa}, \bibinfo{person}{Tore
  Opsahl}, {and} \bibinfo{person}{Kathleen~M Carley}.}
  \bibinfo{year}{2009}\natexlab{}.
\newblock \showarticletitle{Patterns and dynamics of users' behavior and
  interaction: Network analysis of an online community}.
\newblock \bibinfo{journal}{\emph{Journal of the American Society for
  Information Science and Technology}} \bibinfo{volume}{60},
  \bibinfo{number}{5} (\bibinfo{year}{2009}), \bibinfo{pages}{911--932}.
\newblock


\bibitem[Pareja et~al\mbox{.}(2020)]%
        {pareja2020evolvegcn}
\bibfield{author}{\bibinfo{person}{Aldo Pareja}, \bibinfo{person}{Giacomo
  Domeniconi}, \bibinfo{person}{Jie Chen}, \bibinfo{person}{Tengfei Ma},
  \bibinfo{person}{Toyotaro Suzumura}, \bibinfo{person}{Hiroki Kanezashi},
  \bibinfo{person}{Tim Kaler}, \bibinfo{person}{Tao Schardl}, {and}
  \bibinfo{person}{Charles Leiserson}.} \bibinfo{year}{2020}\natexlab{}.
\newblock \showarticletitle{Evolvegcn: Evolving graph convolutional networks
  for dynamic graphs}. In \bibinfo{booktitle}{\emph{AAAI}},
  Vol.~\bibinfo{volume}{34}. \bibinfo{pages}{5363--5370}.
\newblock


\bibitem[Paszke et~al\mbox{.}(2019)]%
        {paszke2019pytorch}
\bibfield{author}{\bibinfo{person}{Adam Paszke}, \bibinfo{person}{Sam Gross},
  \bibinfo{person}{Francisco Massa}, \bibinfo{person}{Adam Lerer},
  \bibinfo{person}{James Bradbury}, \bibinfo{person}{Gregory Chanan},
  \bibinfo{person}{Trevor Killeen}, \bibinfo{person}{Zeming Lin},
  \bibinfo{person}{Natalia Gimelshein}, \bibinfo{person}{Luca Antiga},
  {et~al\mbox{.}}} \bibinfo{year}{2019}\natexlab{}.
\newblock \showarticletitle{Pytorch: An imperative style, high-performance deep
  learning library}.
\newblock \bibinfo{journal}{\emph{Advances in neural information processing
  systems}}  \bibinfo{volume}{32} (\bibinfo{year}{2019}).
\newblock


\bibitem[Pedregosa et~al\mbox{.}(2011)]%
        {pedregosa2011scikit}
\bibfield{author}{\bibinfo{person}{Fabian Pedregosa}, \bibinfo{person}{Ga{\"e}l
  Varoquaux}, \bibinfo{person}{Alexandre Gramfort}, \bibinfo{person}{Vincent
  Michel}, \bibinfo{person}{Bertrand Thirion}, \bibinfo{person}{Olivier
  Grisel}, \bibinfo{person}{Mathieu Blondel}, \bibinfo{person}{Peter
  Prettenhofer}, \bibinfo{person}{Ron Weiss}, \bibinfo{person}{Vincent
  Dubourg}, {et~al\mbox{.}}} \bibinfo{year}{2011}\natexlab{}.
\newblock \showarticletitle{Scikit-learn: Machine learning in Python}.
\newblock \bibinfo{journal}{\emph{Journal of Machine Learning Research}}
  \bibinfo{volume}{12} (\bibinfo{year}{2011}), \bibinfo{pages}{2825--2830}.
\newblock


\bibitem[Peng et~al\mbox{.}(2020)]%
        {peng2020spatial}
\bibfield{author}{\bibinfo{person}{Hao Peng}, \bibinfo{person}{Hongfei Wang},
  \bibinfo{person}{Bowen Du}, \bibinfo{person}{Md~Zakirul~Alam Bhuiyan},
  \bibinfo{person}{Hongyuan Ma}, \bibinfo{person}{Jianwei Liu},
  \bibinfo{person}{Lihong Wang}, \bibinfo{person}{Zeyu Yang},
  \bibinfo{person}{Linfeng Du}, \bibinfo{person}{Senzhang Wang},
  {et~al\mbox{.}}} \bibinfo{year}{2020}\natexlab{}.
\newblock \showarticletitle{Spatial temporal incidence dynamic graph neural
  networks for traffic flow forecasting}.
\newblock \bibinfo{journal}{\emph{Information Sciences}}  \bibinfo{volume}{521}
  (\bibinfo{year}{2020}), \bibinfo{pages}{277--290}.
\newblock


\bibitem[Pope et~al\mbox{.}(2019)]%
        {pope2019explainability}
\bibfield{author}{\bibinfo{person}{Phillip~E Pope}, \bibinfo{person}{Soheil
  Kolouri}, \bibinfo{person}{Mohammad Rostami}, \bibinfo{person}{Charles~E
  Martin}, {and} \bibinfo{person}{Heiko Hoffmann}.}
  \bibinfo{year}{2019}\natexlab{}.
\newblock \showarticletitle{Explainability methods for graph convolutional
  neural networks}. In \bibinfo{booktitle}{\emph{CVPR}}.
  \bibinfo{pages}{10772--10781}.
\newblock


\bibitem[Rossi et~al\mbox{.}(2020)]%
        {rossi2020temporal}
\bibfield{author}{\bibinfo{person}{Emanuele Rossi}, \bibinfo{person}{Ben
  Chamberlain}, \bibinfo{person}{Fabrizio Frasca}, \bibinfo{person}{Davide
  Eynard}, \bibinfo{person}{Federico Monti}, {and} \bibinfo{person}{Michael
  Bronstein}.} \bibinfo{year}{2020}\natexlab{}.
\newblock \showarticletitle{Temporal graph networks for deep learning on
  dynamic graphs}.
\newblock \bibinfo{journal}{\emph{arXiv preprint arXiv:2006.10637}}
  (\bibinfo{year}{2020}).
\newblock


\bibitem[Rudin et~al\mbox{.}(2022)]%
        {ranshomon2}
\bibfield{author}{\bibinfo{person}{Cynthia Rudin}, \bibinfo{person}{Chaofan
  Chen}, \bibinfo{person}{Zhi Chen}, \bibinfo{person}{Haiyang Huang},
  \bibinfo{person}{Lesia Semenova}, {and} \bibinfo{person}{Chudi Zhong}.}
  \bibinfo{year}{2022}\natexlab{}.
\newblock \showarticletitle{Interpretable machine learning: Fundamental
  principles and 10 grand challenges}.
\newblock \bibinfo{journal}{\emph{Statistics Surveys}}  \bibinfo{volume}{16}
  (\bibinfo{year}{2022}).
\newblock


\bibitem[Schlichtkrull et~al\mbox{.}(2020)]%
        {schlichtkrull2020interpreting}
\bibfield{author}{\bibinfo{person}{Michael~Sejr Schlichtkrull},
  \bibinfo{person}{Nicola De~Cao}, {and} \bibinfo{person}{Ivan Titov}.}
  \bibinfo{year}{2020}\natexlab{}.
\newblock \showarticletitle{Interpreting graph neural networks for nlp with
  differentiable edge masking}.
\newblock \bibinfo{journal}{\emph{Preprint arXiv:2010.00577}}
  (\bibinfo{year}{2020}).
\newblock


\bibitem[Seo et~al\mbox{.}(2018)]%
        {seo2018structured}
\bibfield{author}{\bibinfo{person}{Youngjoo Seo}, \bibinfo{person}{Micha{\"e}l
  Defferrard}, \bibinfo{person}{Pierre Vandergheynst}, {and}
  \bibinfo{person}{Xavier Bresson}.} \bibinfo{year}{2018}\natexlab{}.
\newblock \showarticletitle{Structured sequence modeling with graph
  convolutional recurrent networks}. In \bibinfo{booktitle}{\emph{ICONIP}}.
  Springer, \bibinfo{pages}{362--373}.
\newblock


\bibitem[Tonekaboni et~al\mbox{.}(2021)]%
        {tonekaboni2021unsupervised}
\bibfield{author}{\bibinfo{person}{Sana Tonekaboni}, \bibinfo{person}{Danny
  Eytan}, {and} \bibinfo{person}{Anna Goldenberg}.}
  \bibinfo{year}{2021}\natexlab{}.
\newblock \showarticletitle{Unsupervised Representation Learning for Time
  Series with Temporal Neighborhood Coding}. In
  \bibinfo{booktitle}{\emph{ICLR}}.
\newblock


\bibitem[Vaswani et~al\mbox{.}(2017)]%
        {vaswani2017attention}
\bibfield{author}{\bibinfo{person}{Ashish Vaswani}, \bibinfo{person}{Noam
  Shazeer}, \bibinfo{person}{Niki Parmar}, \bibinfo{person}{Jakob Uszkoreit},
  \bibinfo{person}{Llion Jones}, \bibinfo{person}{Aidan~N Gomez},
  \bibinfo{person}{{\L}ukasz Kaiser}, {and} \bibinfo{person}{Illia
  Polosukhin}.} \bibinfo{year}{2017}\natexlab{}.
\newblock \showarticletitle{Attention is all you need}.
\newblock \bibinfo{journal}{\emph{Advances in neural information processing
  systems}}  \bibinfo{volume}{30} (\bibinfo{year}{2017}).
\newblock


\bibitem[Veli{\v{c}}kovi{\'c} et~al\mbox{.}(2017)]%
        {velivckovic2017graph}
\bibfield{author}{\bibinfo{person}{Petar Veli{\v{c}}kovi{\'c}},
  \bibinfo{person}{Guillem Cucurull}, \bibinfo{person}{Arantxa Casanova},
  \bibinfo{person}{Adriana Romero}, \bibinfo{person}{Pietro Lio}, {and}
  \bibinfo{person}{Yoshua Bengio}.} \bibinfo{year}{2017}\natexlab{}.
\newblock \showarticletitle{Graph attention networks}.
\newblock \bibinfo{journal}{\emph{Preprint arXiv:1710.10903}}
  (\bibinfo{year}{2017}).
\newblock


\bibitem[Vu and Thai(2020)]%
        {vu2020pgm}
\bibfield{author}{\bibinfo{person}{Minh Vu} {and} \bibinfo{person}{My~T Thai}.}
  \bibinfo{year}{2020}\natexlab{}.
\newblock \showarticletitle{Pgm-explainer: Probabilistic graphical model
  explanations for graph neural networks}.
\newblock \bibinfo{journal}{\emph{NeurIPS}}  \bibinfo{volume}{33}
  (\bibinfo{year}{2020}), \bibinfo{pages}{12225--12235}.
\newblock


\bibitem[Wang et~al\mbox{.}(2020)]%
        {wang2020traffic}
\bibfield{author}{\bibinfo{person}{Xiaoyang Wang}, \bibinfo{person}{Yao Ma},
  \bibinfo{person}{Yiqi Wang}, \bibinfo{person}{Wei Jin}, \bibinfo{person}{Xin
  Wang}, \bibinfo{person}{Jiliang Tang}, \bibinfo{person}{Caiyan Jia}, {and}
  \bibinfo{person}{Jian Yu}.} \bibinfo{year}{2020}\natexlab{}.
\newblock \showarticletitle{Traffic flow prediction via spatial temporal graph
  neural network}. In \bibinfo{booktitle}{\emph{Proceedings of The Web
  Conference 2020}}. \bibinfo{pages}{1082--1092}.
\newblock


\bibitem[Wu et~al\mbox{.}(2022)]%
        {wu2022graph}
\bibfield{author}{\bibinfo{person}{Shiwen Wu}, \bibinfo{person}{Fei Sun},
  \bibinfo{person}{Wentao Zhang}, \bibinfo{person}{Xu Xie}, {and}
  \bibinfo{person}{Bin Cui}.} \bibinfo{year}{2022}\natexlab{}.
\newblock \showarticletitle{Graph neural networks in recommender systems: a
  survey}.
\newblock \bibinfo{journal}{\emph{Comput. Surveys}} \bibinfo{volume}{55},
  \bibinfo{number}{5} (\bibinfo{year}{2022}), \bibinfo{pages}{1--37}.
\newblock


\bibitem[Xie et~al\mbox{.}(2022)]%
        {xie2022explaining}
\bibfield{author}{\bibinfo{person}{Jiaxuan Xie}, \bibinfo{person}{Yezi Liu},
  {and} \bibinfo{person}{Yanning Shen}.} \bibinfo{year}{2022}\natexlab{}.
\newblock \showarticletitle{Explaining Dynamic Graph Neural Networks via
  Relevance Back-propagation}.
\newblock \bibinfo{journal}{\emph{Preprint arXiv:2207.11175}}
  (\bibinfo{year}{2022}).
\newblock


\bibitem[Xin et~al\mbox{.}(2022)]%
        {ranshomon1}
\bibfield{author}{\bibinfo{person}{Rui Xin}, \bibinfo{person}{Chudi Zhong},
  \bibinfo{person}{Zhi Chen}, \bibinfo{person}{Takuya Takagi},
  \bibinfo{person}{Margo Seltzer}, {and} \bibinfo{person}{Cynthia Rudin}.}
  \bibinfo{year}{2022}\natexlab{}.
\newblock \showarticletitle{Exploring the Whole Rashomon Set of Sparse Decision
  Trees}. In \bibinfo{booktitle}{\emph{NeurIPS}}.
\newblock


\bibitem[Ying et~al\mbox{.}(2019)]%
        {ying2019gnnexplainer}
\bibfield{author}{\bibinfo{person}{Zhitao Ying}, \bibinfo{person}{Dylan
  Bourgeois}, \bibinfo{person}{Jiaxuan You}, \bibinfo{person}{Marinka Zitnik},
  {and} \bibinfo{person}{Jure Leskovec}.} \bibinfo{year}{2019}\natexlab{}.
\newblock \showarticletitle{Gnnexplainer: Generating explanations for graph
  neural networks}. In \bibinfo{booktitle}{\emph{NeurIPS}}.
  \bibinfo{pages}{9240--9251}.
\newblock


\bibitem[Ying et~al\mbox{.}(2018)]%
        {ying2018hierarchical}
\bibfield{author}{\bibinfo{person}{Zhitao Ying}, \bibinfo{person}{Jiaxuan You},
  \bibinfo{person}{Christopher Morris}, \bibinfo{person}{Xiang Ren},
  \bibinfo{person}{Will Hamilton}, {and} \bibinfo{person}{Jure Leskovec}.}
  \bibinfo{year}{2018}\natexlab{}.
\newblock \showarticletitle{Hierarchical graph representation learning with
  differentiable pooling}.
\newblock \bibinfo{journal}{\emph{Advances in neural information processing
  systems}}  \bibinfo{volume}{31} (\bibinfo{year}{2018}).
\newblock


\bibitem[You et~al\mbox{.}(2022)]%
        {you2022roland}
\bibfield{author}{\bibinfo{person}{Jiaxuan You}, \bibinfo{person}{Tianyu Du},
  {and} \bibinfo{person}{Jure Leskovec}.} \bibinfo{year}{2022}\natexlab{}.
\newblock \showarticletitle{ROLAND: graph learning framework for dynamic
  graphs}. In \bibinfo{booktitle}{\emph{ACM SIGKDD}}.
  \bibinfo{pages}{2358--2366}.
\newblock


\bibitem[Yu et~al\mbox{.}(2017)]%
        {yu2017spatio}
\bibfield{author}{\bibinfo{person}{Bing Yu}, \bibinfo{person}{Haoteng Yin},
  {and} \bibinfo{person}{Zhanxing Zhu}.} \bibinfo{year}{2017}\natexlab{}.
\newblock \showarticletitle{Spatio-temporal graph convolutional networks: A
  deep learning framework for traffic forecasting}.
\newblock \bibinfo{journal}{\emph{arXiv preprint arXiv:1709.04875}}
  (\bibinfo{year}{2017}).
\newblock


\bibitem[Yuan et~al\mbox{.}(2020)]%
        {yuan2020xgnn}
\bibfield{author}{\bibinfo{person}{Hao Yuan}, \bibinfo{person}{Jiliang Tang},
  \bibinfo{person}{Xia Hu}, {and} \bibinfo{person}{Shuiwang Ji}.}
  \bibinfo{year}{2020}\natexlab{}.
\newblock \showarticletitle{Xgnn: Towards model-level explanations of graph
  neural networks}. In \bibinfo{booktitle}{\emph{ACM SIGKDD}}.
  \bibinfo{pages}{430--438}.
\newblock


\bibitem[Yuan et~al\mbox{.}(2022)]%
        {yuan2022explainability}
\bibfield{author}{\bibinfo{person}{Hao Yuan}, \bibinfo{person}{Haiyang Yu},
  \bibinfo{person}{Shurui Gui}, {and} \bibinfo{person}{Shuiwang Ji}.}
  \bibinfo{year}{2022}\natexlab{}.
\newblock \showarticletitle{Explainability in graph neural networks: A
  taxonomic survey}.
\newblock \bibinfo{journal}{\emph{IEEE Transactions on Pattern Analysis and
  Machine Intelligence}} (\bibinfo{year}{2022}).
\newblock


\bibitem[Yuan et~al\mbox{.}(2021)]%
        {yuan2021explainability}
\bibfield{author}{\bibinfo{person}{Hao Yuan}, \bibinfo{person}{Haiyang Yu},
  \bibinfo{person}{Jie Wang}, \bibinfo{person}{Kang Li}, {and}
  \bibinfo{person}{Shuiwang Ji}.} \bibinfo{year}{2021}\natexlab{}.
\newblock \showarticletitle{On explainability of graph neural networks via
  subgraph explorations}. In \bibinfo{booktitle}{\emph{ICML}}. PMLR,
  \bibinfo{pages}{12241--12252}.
\newblock


\bibitem[Zhang and Chen(2018)]%
        {zhang2018link}
\bibfield{author}{\bibinfo{person}{Muhan Zhang} {and} \bibinfo{person}{Yixin
  Chen}.} \bibinfo{year}{2018}\natexlab{}.
\newblock \showarticletitle{Link prediction based on graph neural networks}.
\newblock \bibinfo{journal}{\emph{NeurIPS}}  \bibinfo{volume}{31}
  (\bibinfo{year}{2018}).
\newblock


\bibitem[Zhao et~al\mbox{.}(2019)]%
        {zhao2019t}
\bibfield{author}{\bibinfo{person}{Ling Zhao}, \bibinfo{person}{Yujiao Song},
  \bibinfo{person}{Chao Zhang}, \bibinfo{person}{Yu Liu}, \bibinfo{person}{Pu
  Wang}, \bibinfo{person}{Tao Lin}, \bibinfo{person}{Min Deng}, {and}
  \bibinfo{person}{Haifeng Li}.} \bibinfo{year}{2019}\natexlab{}.
\newblock \showarticletitle{T-gcn: A temporal graph convolutional network for
  traffic prediction}.
\newblock \bibinfo{journal}{\emph{IEEE Transactions on Intelligent
  Transportation Systems}} \bibinfo{volume}{21}, \bibinfo{number}{9}
  (\bibinfo{year}{2019}), \bibinfo{pages}{3848--3858}.
\newblock


\end{thebibliography}

\end{document}